\documentclass[lettersize,journal]{IEEEtran}
\usepackage{amsmath,amsfonts}
\usepackage{algorithmic}
\usepackage{algorithm}
\usepackage{array}
\usepackage[caption=false,font=normalsize,labelfont=sf,textfont=sf]{subfig}
\usepackage{textcomp}
\usepackage{stfloats}
\usepackage{url}
\usepackage{verbatim}
\usepackage{graphicx}
\usepackage{cite}
\usepackage{amsthm}
\usepackage{amssymb}
\usepackage{amsmath,lipsum}
\usepackage{epstopdf}
\usepackage{bm}
\usepackage{cuted}%%\stripsep-3pt
\DeclareMathOperator*{\argmin}{argmin}
\DeclareMathOperator*{\argmax}{argmax}
\newcounter{MYtempeqncnt}
\begin{document}

\title{Revisiting Communication-Efficient Federated Learning with Balanced Global and Local Updates}

\author{Zhigang Yan, Dong Li~\IEEEmembership{Senior Member, IEEE, }Zhichao Zhang~\IEEEmembership{Member, IEEE} and Jiguang He

        % <-this % stops a space
\thanks{Zhigang Yan and Dong Li are with the Faculty of Information Technology, Macau University of Science and Technology, Macau, China. (e-mail: 2009853xim20001@student.must.edu.mo and dli@must.edu.mo).

Zhichao Zhang is with the School of Mathematics and Statistics, Nanjing University of Information Science and Technology, Nanjing, China and also with the Faculty of Information Technology, Macau University of Science and Technology, Macau, China (e-mail: zzc910731@163.com).

Jiguang He is with the Technology Innovation Institute, 9639 Masdar City, Abu Dhabi, United Arab Emirates and also with Centre for Wireless Communications, FI-90014, University of Oulu, Finland (e-mail: jiguang.he@oulu.fi).

}

% <-this % stops a space
}



\maketitle

\begin{abstract}
In federated learning (FL), a number of devices train their local models and upload the corresponding parameters or gradients to the base station (BS) to update the global model while protecting their data privacy. However, due to the limited computation and communication resources, the number of local trainings (a.k.a. local update) and that of aggregations (a.k.a. global update) need to be carefully chosen. In this paper, we investigate and analyze the optimal trade-off between the number of local trainings and that of global aggregations to speed up the convergence and enhance the prediction accuracy over the existing works. Our goal is to minimize the global loss function under both the delay and the energy consumption constraints. In order to make the optimization problem tractable, we derive a new and tight upper bound on the loss function, which allows us to obtain closed-form expressions for the number of local trainings and that of global aggregations. Simulation results show that our proposed scheme can achieve a better performance in terms of the prediction accuracy, and converge much faster than the 
baseline schemes.
\end{abstract}


\section{Introduction}
\IEEEPARstart{W}{ith} the rapid development of wireless applications, a huge amount of data is generated every day. One traditional way to deal with these data is to simply send their data to a central controller for processing\cite{ref1}. However, it incurs a new problem for heterogeneous data from different users concerning the privacy requirement or security/legal risks. In recent years, federated learning (FL) has been proposed and emerged as a promising solution to cope with the above challenges\cite{ref2,ref3}. In FL, a group of users train their local models with their own data, and send the training results, e.g., parameters or gradients, to the base station (BS) after several rounds of local training. The BS then aggregates their results to update the global model, and feeds the global parameters or gradients back to all the users to facilitate their local model update. Thus, the entire training process of FL constitutes several rounds of local training and global aggregation. However, the application of FL to wireless networks still faces several challenges, including the non-independent and identically distributed  (non-i.i.d.) data from heterogeneous sources  \cite{ref4,ref5,ref6,ref7,ref8}, data privacy during communication period \cite{ref9,ref10,ref11,ref12,ref13,ref14}, user scheduling for global aggregation \cite{ref15,ref16,ref17,ref18}, and asynchronous local update \cite{ref19,ref20,ref21}. 

Besides the above challenges, the practical implementation of FL suffers from limited communication resources. The existing works can be generally classified into two categories: 1) minimizing the cost/budget to save communication resources \cite{ref22,ref23,ref24,ref25,ref26,ref27,ref28,ref29}, 2) minimizing the loss function subject to communication resources restriction \cite{ref30,ref31,ref32,ref33,ref34}. In the first category, the convergence time in FL with optimal bandwidth allocation and scheduling strategies was studied in \cite{ref22}. Similarly, the delay in FL with optimal scheduling strategies was studied in \cite{ref23}. The minimized delay in FL with optimal bandwidth allocation, scheduling strategies, and transmit power was studied in \cite{ref24}. However, the optimal solutions were only evaluated numerically in~\cite{ref22,ref23}, and there are no closed-form expressions to facilitate the analyses. In addition, the energy cost was not involved in the above works. The total energy consumption was minimized with the latency constraint in~\cite{ref25,ref26}.  Furthermore, minimizing the sum of time and energy cost with the random scheduling strategy was respectively studied in \cite{ref27}, \cite{ref28}, and \cite{ref29}. The optimal number of scheduled users, communication rounds, and training iterations were obtained in~\cite{ref27}. The optimal central processing unit (CPU) frequency, transmit power, and bandwidth allocation were obtained in \cite{ref29}. However, the optimal scheduling strategies with limited resource blocks are only obtained in \cite{ref27,ref28}. 

In the second category, in order to minimize the loss function of FL, the optimal power allocation and scheduling strategy was studied in \cite{ref30}. In addition, the optimal transmit power, resource block allocation, and user selection were investigated in \cite{ref31}. However, both of them did not consider the trade-off between the number of local trainings and that of aggregations. An adaptive FL algorithm to minimize the loss function with optimal number of local trainings and aggregations was proposed in \cite{ref32}. This algorithm was extended in ~\cite{ref33,ref34} to obtain optimal number of scheduled users with resources limitation and total delay cost. 

However, in most of existing works regarding the balanced global and local updates \cite{ref32,ref33,ref34}, there is a need to solve an optimization sub-problem according to the method proposed in \cite{ref32} for each global aggregation, which will slow down the convergence. Besides, it is difficult, if not impossible, to obtain an accurate solution for the sub-problem, resulting in the deteriorated prediction accuracy. These observations motivate our work, and we attempt to break these bottlenecks to avoid solving the sub-problem by obtaining the closed-form expression for the optimal trade-off. Specifically, in this paper, we investigate and analyze the optimal trade-off between the local training and the global aggregation in the FL framework with communication resources restriction. The objective is to minimize the loss function of the global model while satisfying both the delay and energy consumption constraints. Our main contributions are summarized as follows:  

\begin{itemize}
\item{In order to make the optimization problem tractable, we derive a new upper bound of the loss function in a closed-from expression, which can be easily computed and extended to different types of training models and datasets. By doing so, we can avoid computing the optimization sub-problem as in most of existing works, and can thus speed up the convergence and boost the prediction accuracy accordingly. Besides, we also show that the derived upper bound is tight under certain conditions.}

\item{Based on the derived upper bound, we are able to reformulate the original problem, which allows us to obtain closed-from expressions for the number of local trainings and the number of global aggregations. The advantages are that the resultant computational complexity is low, and the impact of the major system parameters can be easily revealed.}

\item{Simulation results show that the proposed scheme can not only significantly improve the convergence speed, but also achieve a better performance than the existing ones in terms of the prediction accuracy.}
\end{itemize}

The remainder of this paper is organized as follows. In Section II, we introduce the system model and formulate the optimization problem. The analysis on the convergence bound is presented in Section III. The convergence analysis is finished in Section IV. We solve the problem by obtaining closed-from expressions for optimized solutions in Section V. The simulation results are shown in Section VI and the concluding remarks are given in Section VII.

\section{System Model and Problem Formulation}

\subsection{Global and local updates for Federated Learning}

\begin{table}[!t]
\caption{List of main notations\label{tab:table1}}
\centering
\begin{tabular}{cc}
\hline
\bf{Notation} & \bf{Definition}\\
\hline
$D_i, D$ & Dataset size of the $i$th user and that of all users\\
$N$ & Number of users\\
$F(\mathbf{w})$, $F_i(\mathbf{w})$ & Loss function of global model and that of the $i$th user model\\
$\mathbf{w}^*$ & Optimal model parameter that minimizes $F(\mathbf{w})$\\ 
$t$,$T$ & Index and total number of local trainings\\
$\mathbf{w}_i(t)$ & Local model parameter of the $i$th user in the $t$th round\\
$\mathbf{w}(t)$ & Global model parameter in the $t$th round\\
$\Tilde{\mathbf{w}}_i(t)$ &  Parameter of the $i$th user after aggregation in the $t$th round\\
$\eta$ & Learning rate\\
$k$,$K$ & Index and total number of aggressions\\
$\tau$ & Number of local trainings between two adjacent aggressions\\
$\tau_{max}$ & Maximum value for $\tau$\\
$\rho$, $\beta$ & Lipschitz parameter and smoothness parameter of $F(\mathbf{w})$\\
$\delta_i$, $\delta$ & Gradient divergence of $\nabla F_i(\mathbf{w})$ and $\nabla F(\mathbf{w})$\\
$P_{tr}$, $P_{cm}$ & Power for local training and communication\\
$t_{tr}$, $t_{cm}$ & Delay of local training and communication\\
$E_{tr}$, $E_{cm}$ & Energy consumption of local training and communication\\
$E_{tot}$, $t_{tot}$ & Maximum energy and delay\\
$B$, $N_{0}$ & System bandwidth and noise power\\
$h_{i,k}$ & channel gain of $i$th user in $k$th aggregation\\
$\mu_{i}$, $a_i$ & fluctuation and maximum of the computation capabilities\\
$Z(\mathbf{w})$ & the number of bits of $\mathbf{w}$ \\
\hline
\end{tabular}
\end{table}

The system under consideration consists of a BS for global aggregation, and $N$ users for local training. For FL, the loss function is commonly utilized for performance evaluation, and the goal is to find the optimal parameter that minimizes the loss function, i.e.,
\begin{equation}
\label{eq1}
\mathbf{w}^* = \argmin\limits_{\mathbf{w}\in \mathbb{R}^n} F(\mathbf{w}),
\end{equation}
where the notations are defined in Table I,\footnote{Note that a bold lowercase letter denotes a vector, $\|\cdot\|$ denotes the Euclidean norm of a vector, $ \mathbb{E}(\cdot) $ denotes a mathematical expectation of variable and $ \mathbb{P}(\cdot) $ denotes the probability of event.} and some examples of loss functions are provided in Table II.\footnote{In the classification model, $y$ and $p$ denote the ground-truth label of a sample and the probability that this model predicts it correctly. While, in the regression model, $\mathbf{x}_i$ is the value vector of features, and $y_i$ is the ground-truth label. $f(\mathbf{x}_i)$ is the predicted value of this model.}

\begin{table}[!t]
\caption{List of loss functions\label{tab:table2}}
\centering
\begin{tabular}{cc}
\hline
\bf{Loss function} & \bf{Classification model}\\
\hline
Log Loss & $F(p)=-(y\log(p)+(1-y)\log(1-p))$\\
Focal Loss & $F(p)=-(1-p)^\gamma \log(p)$\\
Relative Entropy & $F(p)=-y\log(p)$\\
Adaboost & $F(p)=e^{-yp}$\\ 
\hline
\bf{Loss function} & \bf{Regression model}\\
\hline
MSE & $\frac{1}{n}\sum{(y_i-f(\mathbf{x}_i))^2}$\\
MAE & $\frac{1}{n}\sum{|y_i-f(\mathbf{x}_i)|}$\\
\hline
\end{tabular}
\end{table}

In order to obtain $\mathbf{w}^*$ in \eqref{eq1}, it is necessary to conduct local training at each user by using its own data. That is, every user uses its own data to train a local model. Specifically, the $i$th user updates the parameters of its local model by using the following gradient descent method:
\begin{equation}
\label{eq2}
\mathbf{w}_{{i}}({t}) = \mathbf{w}_{{i}}({t}-\mathrm{1})-\eta\nabla {F_i}(\mathbf{w}_{{i}}({t}-\mathrm{1})).
\end{equation}

After several rounds of local training, the local parameters from all the users will be delivered to the BS, which aggregates them to update the global model. In particular, between two adjacent global  aggregations, there will be several rounds of local training. In the paper, we adopt the FedAvg algorithm for the global aggregation \cite{ref3}, which is given by
\begin{equation}
\label{eq3}
\mathbf{w}({t}) = \frac{\sum\limits_{i=\mathrm{1}}^{N}{{D_i}\mathbf{w}_{{i}}({t})}}{{D}}.
\end{equation}
After updating the global model, the BS broadcasts the global parameters to every user. Furthermore, the parameter of the local model  $\Tilde{\mathbf{w}}_{i}(t)$ can be summarized as follow:
\begin{equation}
\label{eq4}
{\Tilde{\mathbf{w}}_{i}(t)} = \begin{cases}
\frac{\sum\limits_{i=1}^N{{D_i}\mathbf{w}_{{i}}({t})}}{{D}},& t=k\tau, \\ 
{\mathbf{w}_{{i}}({t}-\mathrm{1})-\eta\nabla {F_i}(\mathbf{w}_{{i}}({t}-\mathrm{1})),}&{\text{otherwise}}. 
\end{cases}
\end{equation}
The parameter of the local model is updated by using \eqref{eq3} when $t=k\tau$. Otherwise, its update follows \eqref{eq2}.

The global loss function is defined as\footnote{The index $(t)$ is omitted thereafter for simplicity of illustration.}
\begin{equation}
\label{eq5}
{F}(\mathbf{w})=\frac{\sum\limits_{{i}=\mathrm{1}}^{{N}} {D_i F_i}(\mathbf{w})}{{D}}.
\end{equation}
It is indicated that ${F}(\mathbf{w})$ cannot
be calculated directly by following Table II due to the lack of users' data at the BS. Instead, we refer to \eqref{eq5} for computing ${F}(\mathbf{w})$ after global aggregation.

\subsection{System Delay Model}

\textit{1) Local Training Delay: }According to \cite{ref33} and \cite{ref35}, the local training delay of the $i$th user (e.g., $t_{i,tr}$) follows the shifted exponential distribution, it means that

\begin{equation}
\label{eq6}
{\mathbb{P}(t_{i,tr}-a_{i}\tau d_{i}\leq t)} = \begin{cases}
1-e^{-\frac{\mu_{i}}{\tau d_{i}}t},& t\geq 0, \\ 
{0,}&{\text{otherwise}}, 
\end{cases}
\end{equation}
where $d_i$ is the batch size of local training of the $i$th user. If we apply the Stochastic Gradient Descent to update all local models, we simply set $d_i=1$ to any $i$. Besides, $\mu_i$ and $a_i$ denote the fluctuation and the maximum of the computation capabilities \cite{ref33}. If $d_i$ is the same, the $a_i$ will be the same \cite{ref35}. To this end, we consider $\mu_i=\mu$ to any $i$.

In order to achieve synchronous update of all local models, all users are allowed to send parameters to the BS when all of them finish the local training. Therefore, the training delay of all users (e.g., $t_{tr}$) depends on the maximal $t_{i,tr}$, namely,  $t_{tr}\triangleq\max \{t_{i,tr}\}$.

\textit{Theorem 1: The upper bound of mathematical expectation of $t_{tr}$ is given by}

\begin{equation}
\label{eq7}
\mathbb{E}(\max\{t_{i,tr}\})\leq\frac{N\tau}{\mu}\cdot I_{0}+a\tau,
\end{equation}
where $a=\max\{a_i\}$, and $I_{0}=\sum_{i=1}^{N}\frac{C_{N-1}^{i-1}(-1)^{i-1}}{i^2}$.

\textit{Proof: See Appendix A.$\hfill\qedsymbol$}

Theorem 1 shows that, a higher frequency of aggregation will lead to a lower average local training delay.

\textit{2) Communication Delay: }After several rounds of local training, all users need to send their local parameters to the BS. Then the BS averages the parameters and broadcasts to all users. Therefore, the communication delay includes the upload and download delay. However, we ignore the download delay in this paper because the transmit power of BS is large enough to make the download delay small.

The upload delay depends on the bit number of parameters which will be sent (e.g., $Z(\mathbf{w})$) and the achievable transmission rate of the system. We apply an FDMA system in the communication part of FL, therefore, the communication delay is

\begin{equation}
\label{eq8}
t_{i,cm}=\frac{Z(\mathbf{w})}{B\log_{2}(1+\frac{P_{cm}h_i}{N_{0}})},
\end{equation}
where $B$ and $h_{i}$ is the bandwidth and the channel gain of the $i$th user respectively. $N_{0}$ is the noise power. In our communication delay model, $t_{i,cm}$ will be treated as a constant. This treatment is common in works similar to this paper, such as \cite{ref33} and \cite{ref34}. Besides, similar to the analysis of the local training delay, we consider the communication delay of all users denoted as $t_{cm}=\max\{t_{i,cm}\}$.

\subsection{System Energy Consumption Model}

\textit{1) Energy consumption of local training: }The energy consumption of local training depends on the computation capacity of the user. According to \cite{ref25}, the energy consumption of one round of local training in the $i$th user can be written as

\begin{equation}
\label{eq9}
E_{i,tr}=\kappa CD_{i}a^{2}_{i},
\end{equation}
where $\kappa$ is the effective switched capacitance that depends on the chip architecture. $C$ is is the number of CPU cycles, and $E_{tr}=\max\{E_{i,tr}\}=\kappa CD_{max}a^{2}$.

\textit{2) Energy consumption of communication: } We have obtain the communication delay in \eqref{eq8}, therefore, the Energy consumption of communication is $E_{cm}=P_{cm}t_{cm}$, and $P_{cm}$ will be also treated as a constant  \cite{ref25}.

\subsection{Problem Formulation}
In this paper, our goal is to find the optimal trade-off between the number of local trainings and that of global aggregations by minimizing the global loss function. The optimization problem is formulated as
\begin{align}
&\min_{T,K}\quad {F}(\mathbf{w}) \label{eq10}\\
&\;\textrm{s.t.}\quad \mathbb{E}(t_{tr})\cdot T+t_{cm}K\leq t_{tot}, \tag{\ref{eq10}{a}} \label{10a}\\
&\quad\quad E_{tr}\cdot T+E_{cm}K\leq E_{tot}, \tag{\ref{eq10}{b}}\label{10b}\\
&\quad\quad t\leq T,\tag{\ref{eq10}{c}}\label{10c}\\
&\quad\quad T=K\tau,\tag{\ref{eq10}{d}}\label{10d}
\end{align}
where (10a) and (10b) denote the delay and energy consumption constraints, respectively. It is noted that in both (10a) and (10b), we focus on the communication resources for the local model and neglect those for the global model similar to existing works. This is due to the obvious contrast between BS and users in terms of the computation capacity and the energy availability. Besides, (10c) and (10d) require that the number of the global aggregations should be less than or equal to the total number of local trainings, and the total number of local trainings should be an integer-fold of the number of aggregations.

\section{Convergence Analysis of FL}

The optimization problem (10) has optimal solution means the loss fuction of FL is convergent. Therefore, we are going to analyze convergence of the loss function and its convergence rate in this section.

The following assumptions are utilized to finish the convergence analysis.

\subsubsection*{ Assumption 1}

\textit{For each user and any pair of $\mathbf{w}$ and $\mathbf{w}'$}

\begin{enumerate}
\item ${F_i}(\mathbf{w})${ is convex}
\item ${F_i}(\mathbf{w})${ is $\rho$-Lipscitz: $\Vert {F_i}(\mathbf{w}) - {F_i}(\mathbf{w}') \Vert \leq \rho \Vert \mathbf{w}-\mathbf{w}' \Vert $}
\item ${F_i}(\mathbf{w})${ is $\beta$-Smooth: $\Vert \nabla{F_i}(\mathbf{w}) - \nabla{F_i}(\mathbf{w}') \Vert \leq \beta \Vert \mathbf{w}-\mathbf{w}' \Vert $}

\item \textit{ (Gradient divergence) For any $i$ and $\mathbf{w}$, $\delta_i$ is an upper bound of $\Vert\nabla{F_i}(\mathbf{w}) - \nabla{F}(\mathbf{w})\Vert$, i.e.,}

\begin{equation}
\label{eq11}
\Vert\nabla{F_i}(\mathbf{w}) - \nabla{F}(\mathbf{w})\Vert \leq \delta_{{i}}.
\end{equation}

We define $\delta=\frac{\sum_{i=1}^{N}D_i\delta_i}{D}$.

\item \textit{ (Federated learning gap) $\epsilon$ is the lower bound of $ {F}(\mathbf{w}({T}))-{F}(\mathbf{w}^*)$, i.e.,}

\begin{equation}
\label{eq12}
{F}(\mathbf{w}({T}))-{F}(\mathbf{w}^*) \geq \epsilon.
\end{equation}
\end{enumerate}

A proportion of loss functions are shown in Table II satisfying 1)-3) in Assumption 1. These loss functions are often used in some classification models and regression models, such as support vector machine (SVM), convolutional neural network (CNN), and linear regression, which will be examined later in our simulations in Section V. Regarding 4) of Assumption 1, gradient divergence reflects the difference among data distributions of all the users. If the data distributions are similar to each other, the gradient divergence will be small, and vice versa. Furthermore, 5) of Assumption 1 reflects the difference between model trained by FL and that trained by centralized machine learning.

\textit{Assumption 2: The gradient of the global model is upper bounded by}
\begin{equation}
\label{eq13}
\nabla \mathit{F}^* = \argmax\limits_{\nabla\mathit{F}(\mathbf{w})\in \mathrm{\mathbb{R}^n}} \Vert\nabla{F}(\mathbf{w})\Vert.
\end{equation}

Assumption 2 is motivated by the fact that the loss function of the global model can only be calculated based on the information (e.g., data and parameters) about the local model. Meanwhile, the number of aggregations is limited, so the gradients of global loss function are restricted in a finite set. 

According to Assumptions 1 and 2, we are able to propose the following lemmas.

\textit{Lemma 1: ${F}(\mathbf{w})$ is also convex, $\rho$-Lipscitz, and $\beta$-Smooth.}

\textit{Proof: The proof is straightforward based on 1)-3) in Assumption 1, and thus omitted for brevity.$\hfill\qedsymbol$}

\textit{Lemma 2: The gradient of local model is upper bounded by}

\begin{equation}
\label{eq14}
\Vert\nabla{F_i}(\mathbf{w})\Vert \leq \delta_{{i}} + \nabla {F}^*.
\end{equation}

\textit{Proof: See Appendix B.$\hfill\qedsymbol$}

Now, based on Lemmas 1 and 2, we are ready to present the following theorem.

\textit{Theorem 2: For any $i$ and $t$, we have}

\begin{equation}
\label{eq15}
\Vert\Tilde{\mathbf{w}}_{{i}}({t})-\mathbf{w}({T})\Vert \leq g_i(t),
\end{equation}
where $g_i(t) = (\delta_i + \nabla F^*)\eta t - \frac{\tau}{\rho}$. 

\textit{Proof: See Appendix C.$\hfill\qedsymbol$}

Theorem 2 gives an upper bound of the difference between the parameters of the $i$th user after the $t$th local training and the parameters of global model after the $T$th local training. From Theorem 2, it is known that the smaller the upper bound, the better the local model. Recall that $t=k\tau$ when FL aggregates after the $t$th local training. In particular, when $t=0$, $k = 0$ and $g_i(0)=0$. In FL, a larger $\tau$ will cause the local model overfitting. On the contrary, a smaller $\tau$ makes the local model underfitting. In Theorem 1, $k$ should be larger than or equal to $\lceil \frac{1}{\rho\eta(\delta_i+\nabla F^*)}\rceil$, leading to a positive value of $g_i(t)$ in practice.

Based on Theorem 2, we obtain the following theorem. 

\textit{Theorem 3: The upper bound of  ${F}(\mathbf{w}({T}))-{F}(\mathbf{w}^*)$ is given by}

\begin{equation}
\label{eq16}
\mathit{F}(\mathbf{w}(\mathit{T}))-\mathit{F}(\mathbf{w}^*) \leq \frac{\epsilon ^2}{\rho {g_i}(\mathrm{1})+\rho\eta\nabla {F^* T}}.
\end{equation}

\textit{Proof: See Appendix D.$\hfill\qedsymbol$}

It is observed from Theorem 3 that a larger $T$ and a smaller $\tau$ lead to a lower upper bound of ${F}(\mathbf{w}({T}))-{F}(\mathbf{w}^*)$. Besides, it can be also deduced from Theorem 2 that how the distribution of the data influence the convergence. Provided that $\epsilon \approx \delta_i$, the denominator $\rho {g_i(\mathrm{1})}+\rho\eta\nabla {F^* t}$ are much larger than the numerator $\epsilon^2$, so the numerator is more sensitive to value changes of $\epsilon$. For non-i.i.d. data, $\epsilon$ and $\delta_i$ will be larger compared to the case of i.i.d. data. This in turn indicates that the non-i.i.d. data will lead to a larger upper bound, which will be verified in Section V.

Theorem 3 also reflects that, if we fix the frequency of aggregation (e.g., $\tau$), $\frac{\epsilon ^2}{\rho {g_i}(\mathrm{1})+\rho\eta\nabla {F^* t}}$ decreases with increasing rounds of local training (e.g., $t$). Thus, the loss function of FL can be shown to be convergent.

In order to get more insight into Theorem 3, we present two corollaries, which depicts the tightness and the number of local training after convergence, respectively.

\textit{Corollary 1: If the parameters satisfy the following conditions, the upper bound in (16) is $\epsilon$-tight.} 

1) $\mathbf{w}_{i}(1)=m \mathbf{w}(T), m\in\mathbb{R}^{+},\forall i$.

2) $\nabla F_{i}(\mathbf{w}(t))=p \nabla F(\mathbf{w}(t)),p\in\mathbb{R}^{+},\forall i$.

3) $\Tilde{\mathbf{w}}_{i}(t)=q \mathbf{w}(t),q\in\mathbb{R}^{+},\forall i$.

4) $\mathbf{w}(t)-\mathbf{w}(T)=-s\nabla F(\mathbf{w}(t)),s\in\mathbb{R}^{+}$.

\textit{Proof: In the proof of Theorem 3, if the equality condition of each step can be satisfied, the upper bound of Theorem 3 will be tight. That is to say, when conditions 1)-4) are satisfied, (36)-(47) are achievable. Thus, we can arrive at Corollary 1 after some manipulations. $\hfill\qedsymbol$}

In Corollary 1, 1) means that the first parameter vectors of all users are in parallel and with the same direction as the final global parameter vector. 2) means the gradient of every users are in parallel and with the same direction as the global gradient in each round of aggregate. 3) means if one round will aggregate, this parameter vector after aggregation is in parallel and with the same direction as this vector without any aggregation. 4) means the global parameter vector changes in the opposite direction to the gradient. It is not difficult to see that when the FL system has only one user, all conditions will be satisfied. Besides, if the loss function also satisfies that $\epsilon = 0$ which is the infimum of ${F}(\mathbf{w}({T}))-{F}(\mathbf{w}^*)$ and  $\rho=F'(\xi)$, where $\xi$ is the value which satisfies Lagrange's Mean Value Theorem of ${F}(\mathbf{w})$, the upper bound in (16) is $\epsilon$-tight.

From the Theorem 3, we obtain the convergence gap of FL. Thus, we can use the number of local trainings which the gap can arrive a small enough value (e.g., $\varepsilon$) to reflect the convergence rate. Therefore, we obtain the proposition of Theorem 3 as follow:

\textit{Corollary 2: }The number of local trainings after $\varepsilon$ convergence is at least $\frac{1}{\rho\eta\nabla F^*}(\frac{\epsilon^2}{\varepsilon}-\rho\eta(\delta_i + \nabla F^*) + \tau)$.

\textit{Proof: The proof simply follows from (16), and thus omitted for brevity.$\hfill\qedsymbol$}

\section{Optimal Trade-off between the Local Trainings and the Global Aggregations}

In this section, we are going to solve the optimization problem in \eqref{eq10} with constraints (10a)-(10d), which is, however, hard to deal with due to the intractable objective function in \eqref{eq10}. In \cite{ref32}, an adaptive method was proposed to transform \eqref{eq10} to facilitate the optimization process. However, it incurs a new optimization sub-problem in each iteration, which harms the convergence speed. In this paper, we propose a new method to circumvent this problem, in which \eqref{eq10} is derived in a closed-form expression.

Finally, the objective function in \eqref{eq10} can be replaced by the upper bound in \eqref{eq16}, and the original optimization problem is reformulated as

\begin{align}
&\min_{\tau,K}\quad \frac{\epsilon^2}{\rho {g_i(\mathrm{1})}+\rho\eta\nabla {F^* K\tau}},\label{eq17}\\
&\;\textrm{s.t.}\quad (\frac{N}{\mu}\cdot I_{0}+a)K\tau^{2}+t_{cm}K\leq t_{tot}, \tag{\ref{eq17}{a}} \label{17a}\\
&\quad\quad E_{tr}K\tau+E_{cm}K\leq E_{tot}. \tag{\ref{eq17}{b}}\label{17b}
\end{align}

\subsection{Proposed Solution}
In \eqref{eq13}, $\epsilon^2$, $\rho\eta(\delta + \nabla {F^*})$, and $\rho\eta\nabla{F^*}$ are constants, and  $\rho\eta(\delta + \nabla {F^*}) + (\rho\eta\nabla{F^*}-\frac{1}{K})T>\textrm{0}$. In this regard, the optimization problem in \eqref{eq17} can be further transformed into the following:
\begin{align}
&\min_{\tau,K}\quad \tau-\rho\eta\nabla{F^*}K\tau, \label{eq18}\\
&\;\textrm{s.t.}\quad (17\mathrm{a}), (17\mathrm{b}).\nonumber
\end{align}
It can be checked that \eqref{eq18} is non-convex over $K$ and $\tau$, which renders the problem hard to solve. However, the following theorem reveals the optimal solutions to \eqref{eq18}.

\textit{Theorem 4: The solution that satisfies the KKT conditions of \eqref{eq18} is given by \eqref{eq19} on the top of this page, where} $I_{1}=E^{2}_{tr}t^{2}_{tot}-4E_{tot}(I_{0}N\mu+a)(t_{cm}E_{tot}-P_{cm}t_{cm}t_{tot})$. Let $K_i$ and $\tau_i$ as the values of \eqref{eq19}, $i=1,2$.  The optimal solution of 
\eqref{eq18} is given by \eqref{eq20} on the top of this page.

\begin{figure*}[!t]
\normalsize
%\newcounter{MYtempeqncnt}
\setcounter{MYtempeqncnt}{\value{equation}}
\setcounter{equation}{18}
\vspace*{4pt}
\begin{equation}
\label{eq19}
{(\tau_i, K_i)} = \begin{cases}
\Big(\frac{1}{E_{tr}}(\sqrt{\rho\eta\nabla{F^*}P_{cm}t_{cm}E_{tot}}-P_{cm}t_{cm}),  \sqrt{\frac{E_{tot}}{\rho\eta\nabla{F^*}P_{cm}t_{cm}}} \Big), i=1, \\ 
\Big(\frac{E_{tr}t_{tot}-\sqrt{I_{1}}}{2E_{tot}(I_{0}N\mu+a)},\frac{2E^{2}_{tot}(I_{0}N\mu+a)}{2(I_{0}N\mu+a)E_{tot}P_{cm}t_{cm}+E_{tr}(E_{tr}t_{tot}-\sqrt{I_{1}})} \Big), i=2, \\
\end{cases} 
\end{equation}

\begin{equation}
\label{solution}
(\tau^*, K^*)=\begin{cases}
\argmin_{i}~\tau_i-\rho\eta\nabla{F^*}K_i\tau_i,i=1,2, &{\text{if}}~\tau_i, K_i \geq 1, I_1\geq 0,\\
(\tau_1, K_1) &{\text{if}}~\tau_1, K_1 \geq 1, I_1\geq 0, \tau_2, K_2 < 1~{\text{or}}~\tau_1, K_1 \geq 1, I_1< 0\\
(\tau_2, K_2) &{\text{if}}~\tau_2, K_2 \geq 1,I_1\geq 0, \tau_1, K_1 < 1.\\
\end{cases} 
\end{equation}

\hrulefill
\end{figure*}
\setcounter{equation}{\value{MYtempeqncnt}}

\textit{Proof: See Appendix E.$\hfill\qedsymbol$}

From the Theorem 4, it is seen that the optimal number of local trainings and  global aggregations also depend on the data distribution at each user. The i.i.d. data is beneficial for training and thus the loss function is easy to convergence. Therefore, the i.i.d. data brings a higher $\nabla F^*$ and a lower training delay.

\begin{algorithm}
	\renewcommand{\algorithmicrequire}{\textbf{Input:}}
	\renewcommand{\algorithmicensure}{\textbf{Output:}}
	\caption{Proposed Algorithm for Communication-Efficient FL}
	\label{alg}
	\begin{algorithmic}[1]
		\STATE Initialization: $\rho, \eta, \delta_i, \nabla F^*, \tau_{max}$
		\STATE  Initialization: $\mu, a, t_{cm}, P_{tr}, P_{cm}, t_{tot}, E_{tot}$
		\STATE $//$ \textit{Specific values of these hyperparameters are shown in Tables III and IV}.
		\STATE Compute $I_{0}$ by \eqref{eq7}
		\IF{$K^*_i$ satisfies conditions in Theorem 4}
		\STATE Compute the optimal $\tau^*, K^*$ by Theorem 4
		\STATE $T^* \gets K^*\tau^*$
		\STATE $\tau \gets \min\{ \tau^* ,\tau_{max} \}$
		\ENDIF
		\STATE Initiate $\mathbf{w}_{i}(1)$ to the same value for all $i$ 
		\FOR{$t=1,2,\dots, T^*$}
		\STATE For each user $i$ in parallel, finish the local training to update $\mathbf{w}_{{i}}({t})$ 
		\STATE $//$ \textit{Local training}.
        \IF{$t$ is an integer-fold of $\tau$}
        \STATE $\mathbf{w}_{{i}}({t}) \gets \Tilde{\mathbf{w}}_{{i}}({t})$
        \STATE $//$ \textit{Global aggregation}.
        \ENDIF
		\ENDFOR
		\ENSURE  $\mathbf{w}({T^*})$
	\end{algorithmic}  
\end{algorithm}

In Algorithm 1, lines 1 and 2 provide the initial values to the hyperparameters. $\tau_{max}$ is the maximal $\tau$ to prevent the model overfitting during the training. Line 4 computes the $\lambda$ to determine the optimal value $\tau$ and $K$. Line 8 sets the range value of $\tau$ as $\{1,2,\dots, 20\}$ to avoid overfitting. Lines 10 to 17 are the detailed steps of FL, which include the local training phase and the communication phase. The maximum of $t$ is $T^*$, which is obtained in Line 7. When $t$ is an integer-fold of $\tau$, the FL performs the aggregation. Finally, the algorithm outputs the $\mathbf{w}(T^*)$.

\section{Simulation Results}
\subsection{System Setup}
\textit{1) Parameters:} For our simulations, we consider an FL system with five users and one BS.  For simplicity of illustration, we run simulations to compute the average time delay of local training and communication/aggregation. To be specific, we have $t_{cm}=0.14$s in the SVM model, and $t_{cm}=0.143$s in the CNN model. Because CNN has more parameters than SVM, it has a larger $Z(\mathbf{w})$. Thus its $t_{cm}$ longer is realistic.  The other parameters used in the simulations are listed in Table~III.

\begin{table}[!t]
\caption{System parameters\label{tab:table3}}
\centering
\begin{tabular}{cc||cc}
\hline
Parameter & Value & Parameter & Value\\
\hline
$\eta$ & 0.1 & $a$ & 2GHz\\
$\rho$ & 0.01 & $P_{cm}$ & 1.5W\\
$t_{tot}$ & 200s & $E_{tot}$ & 1500J\\
$E_{tr}$ & 10J & $\mu$  & 0.2 \\
\hline
\end{tabular}
\end{table}

To reflect the lower convergence speed in the case of non-i.i.d. data for model training, we set a lower value of $\nabla F^*$ in the simulation. In general, the CNN model has more parameters than the SVM model, so we set a larger value of $\nabla F^*$ in CNN. Finally, in the same model, training on a larger dataset makes the value of $\nabla F^*$ larger. We summarize $\nabla F^*$ in different cases in Table IV.

\begin{table}[!t]
\caption{$\nabla F^*$ in Different Cases\label{tab:table4}}
\centering
\begin{tabular}{cc}
\hline
Case & Value \\
\hline
SVM + i.i.d. + MNIST & 1000 \\
SVM + non-i.i.d. + MNIST & 350 \\
CNN + i.i.d. + MNIST & 1300 \\
CNN + non-i.i.d. + MNIST & 1000 \\
CNN + i.i.d. + CIFAR-10 & 1400 \\
CNN + non-i.i.d. + CIFAR-10 & 600 \\
\hline
\end{tabular}
\end{table}

\begin{figure*}[!t]
\centering
\subfloat[]{\includegraphics[width=2.1in]{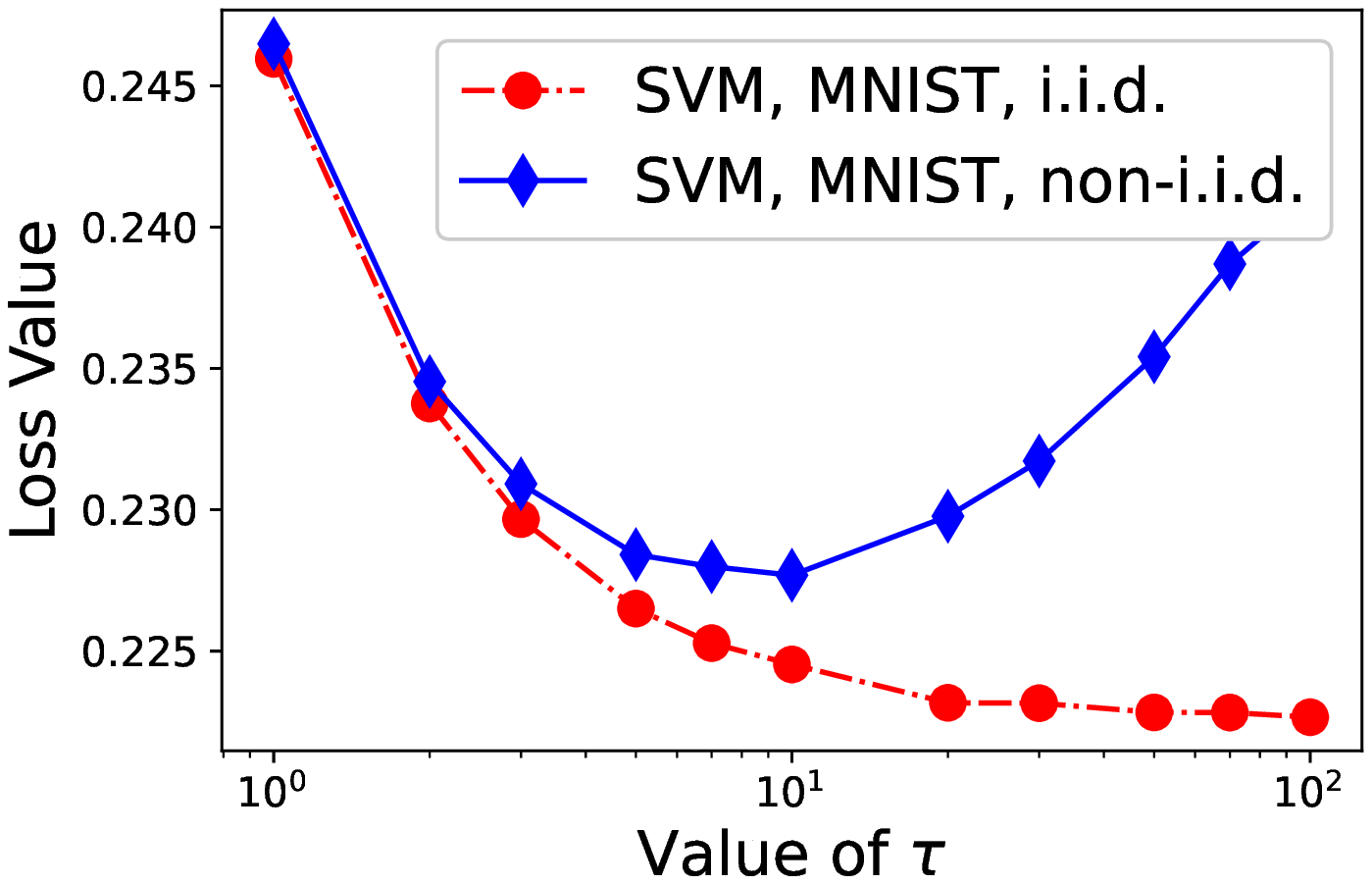}%
\label{fig1_first_case}}
\subfloat[]{\includegraphics[width=2.1in]{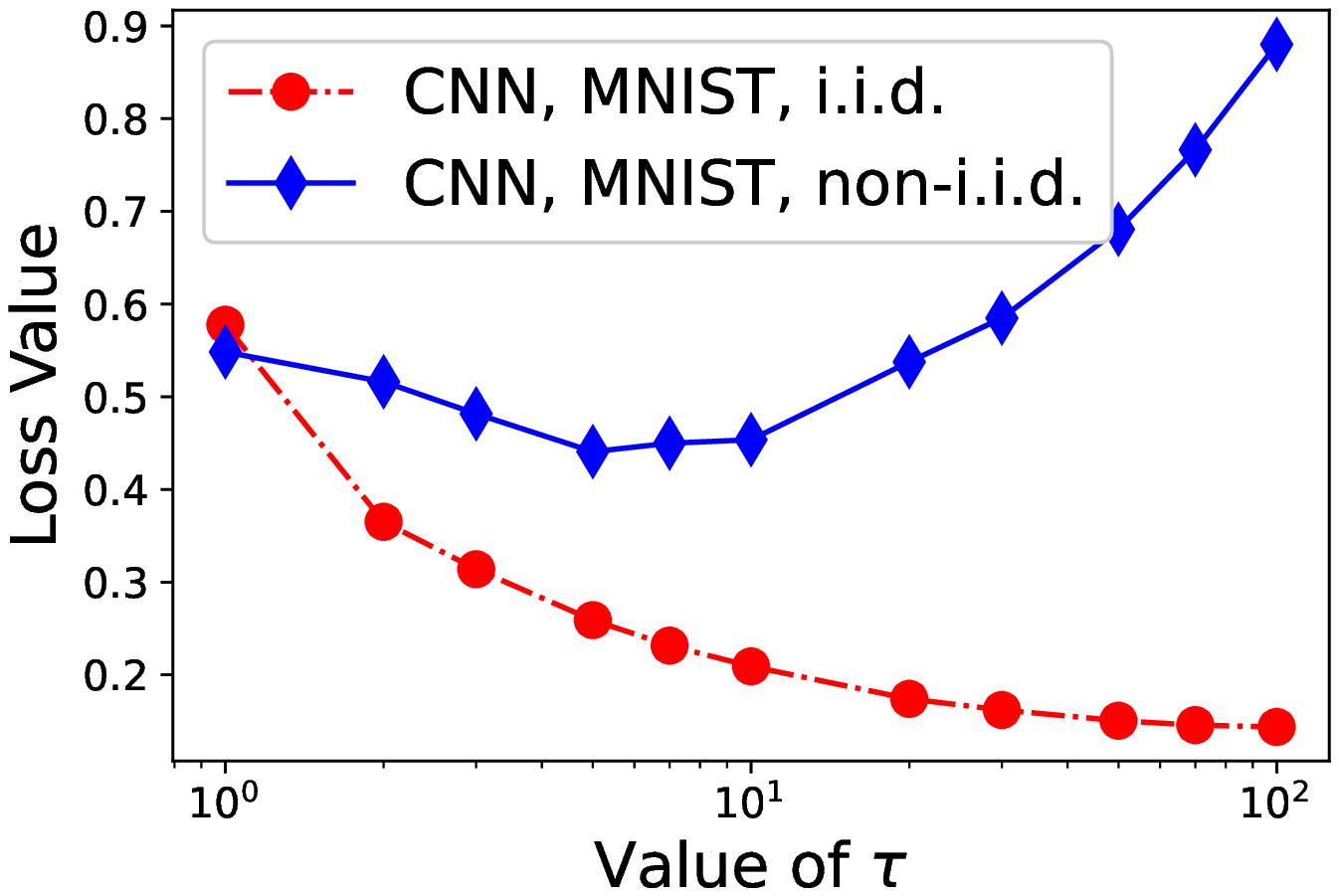}%
\label{fig1_second_case}}
\subfloat[]{\includegraphics[width=2.1in]{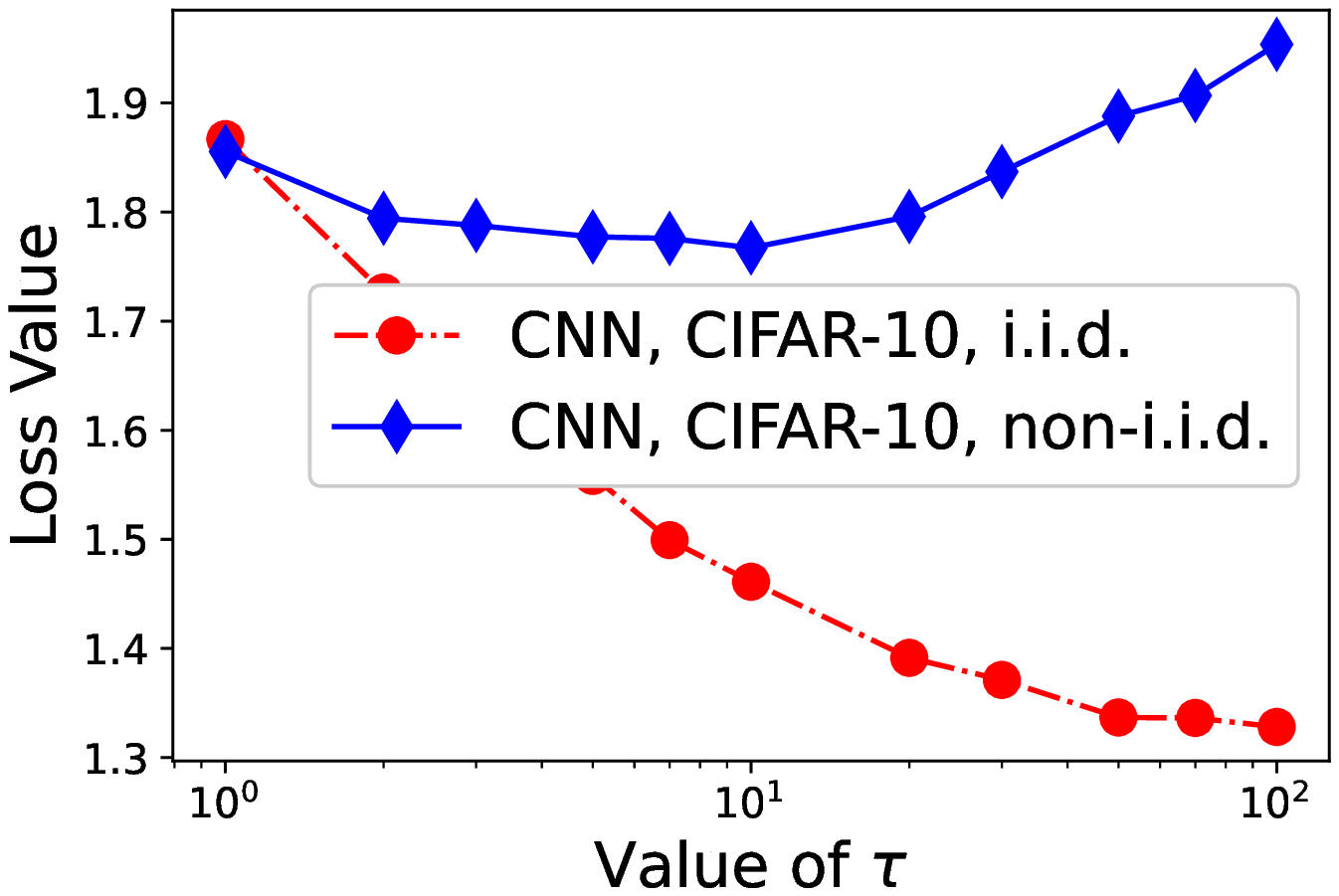}%
\label{fig1_third_case}}
\label{fig_sim}
\subfloat[]{\includegraphics[width=2.1in]{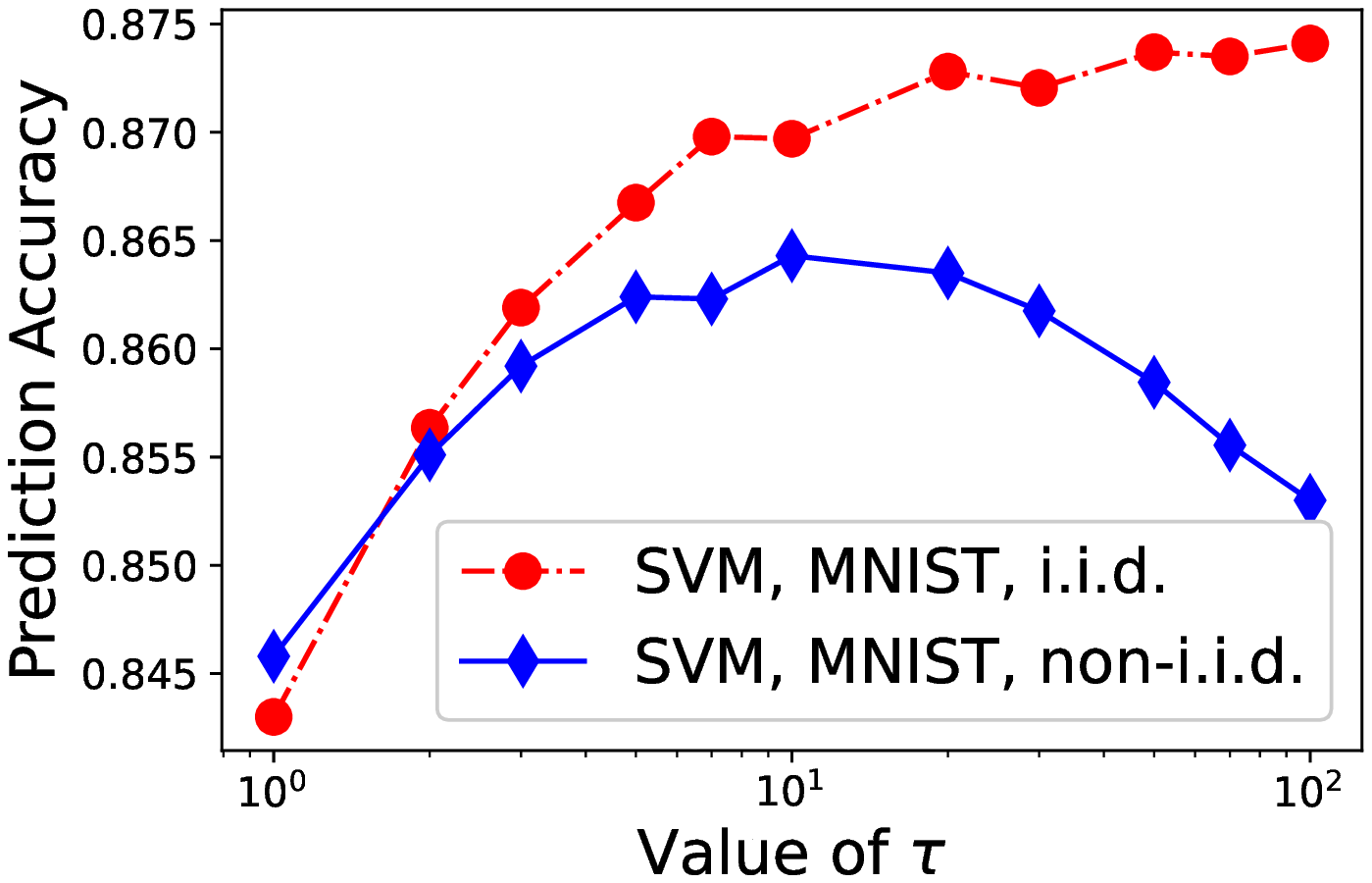}%
\label{fig1_4_case}}
\subfloat[]{\includegraphics[width=2.1in]{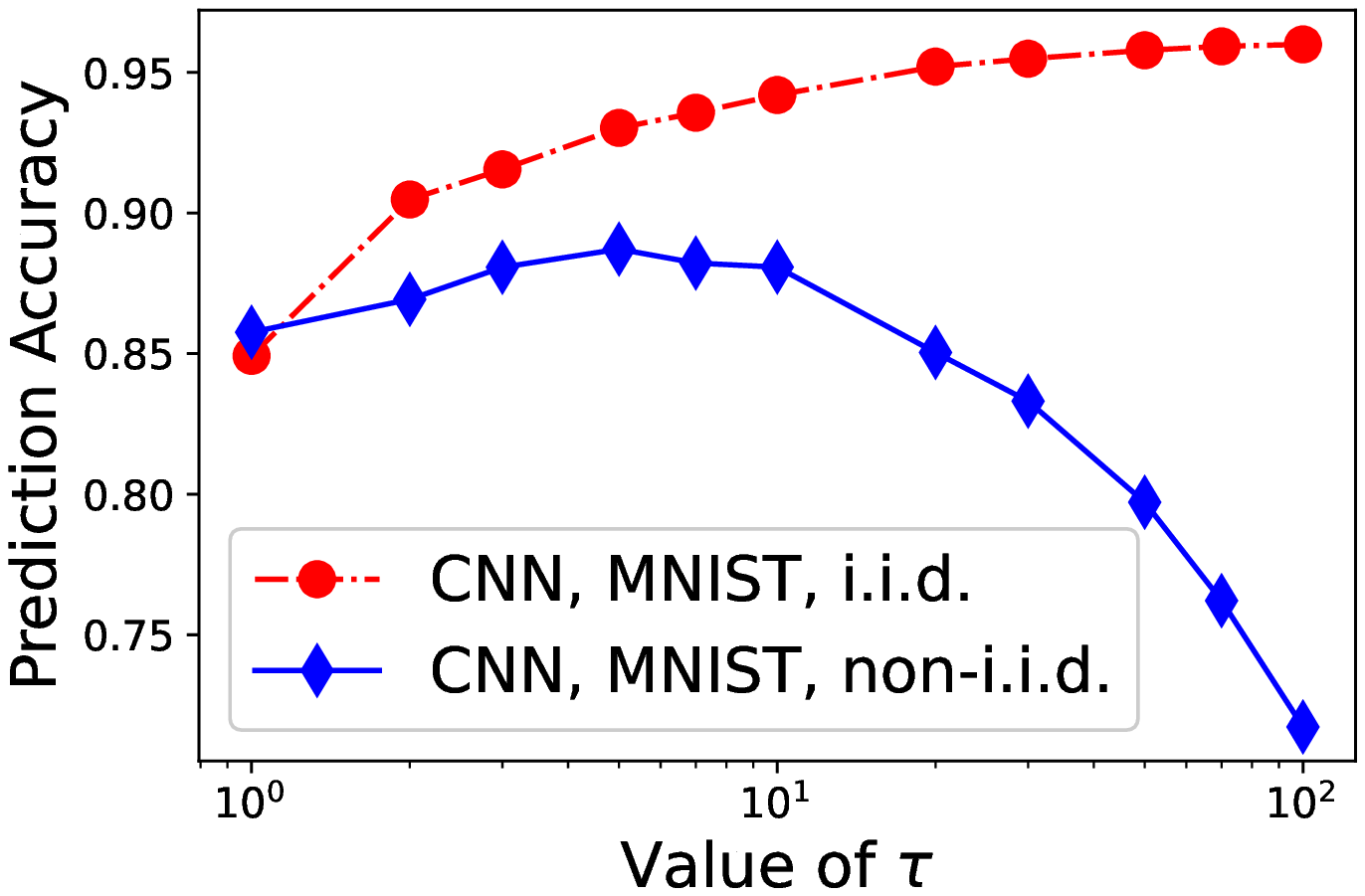}%
\label{fig1_5_case}}
\subfloat[]{\includegraphics[width=2.1in]{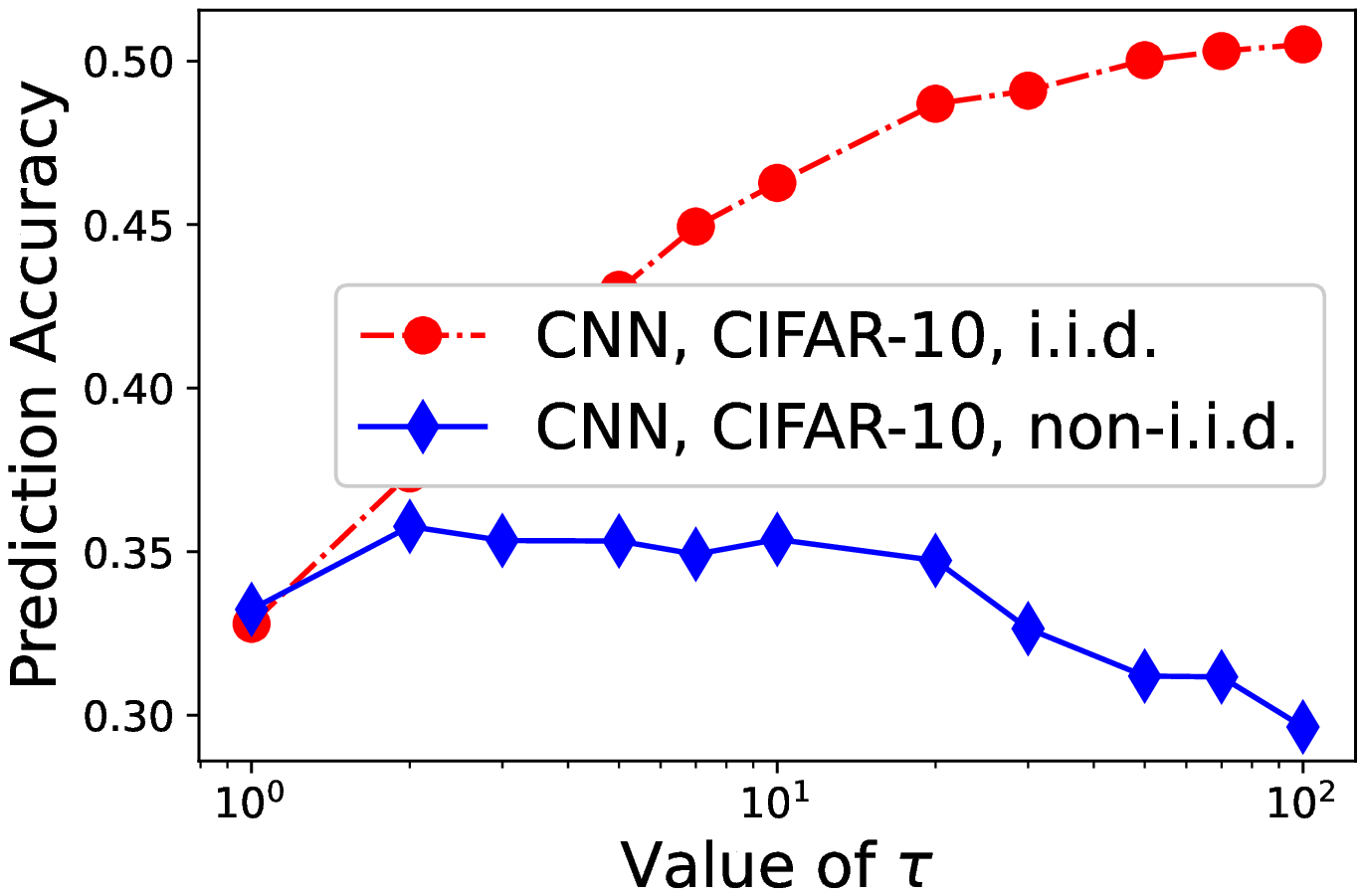}%
\label{fig1_6_case}}

\caption{Impact of the non-i.i.d. datasets with different models, datasets, and $\tau$'s. 
(a) Loss value for SVM with MNIST. (b) Loss value for CNN with MNIST. 
(c) Loss value for CNN with CIFAR-10. (d) Prediction accuracy for SVM with MNIST. 
(e) Prediction accuracy for CNN with MNIST. (f) Prediction accuracy for CNN with CIFAR-10.}
\label{fig1_sim}
\end{figure*}

\textit{2) Datasets:}
In our simulations, for simplicity of  comparison, we consider $4$ different datasets, which represent different data sizes, sample complexities, and data distributions. All the datasets are used for training a classification model. 

The first dataset is MNIST, which contains $70, 000$ gray-scale images of  handwritten digits ($60, 000$ for training and $10, 000$ for testing). The second one is CIFAR-10, which contains $60, 000$ color images ($50, 000$ for training and $10, 000$ for testing) of $10$ different types of objects. Compared to MNIST, CIFAR-10 represents a lower data size but larger number of channels. 

Besides, we consider $2$ cases to make these two datasets have different data distributions in our simulations. In case $1$, each data sample is randomly assigned to a user, so each user has i.i.d. data. In case $2$, all the data samples in each user have the same label. However, the samples in the whole datasets have multiple different labels. Therefore, each user have non-i.i.d. data in case $2$. 

\textit{3) Models:}
We use 2 different models in our simulations for the classification tasks. 
The first one is SVM for binary classification, which is applied in MNIST dataset to predict whether the digit in the sample picture is even or odd. The second one is CNN, which is applied in MNIST dataset for binary classification (same as SVM) and CIFAR-10 dataset for multi-class classification.  

\textit{4) Baseline:} We take the adaptive FL algorithm \cite{ref32} as the baseline for our work.

\begin{figure*}[!t]
\centering
\subfloat[]{\includegraphics[width=3in]{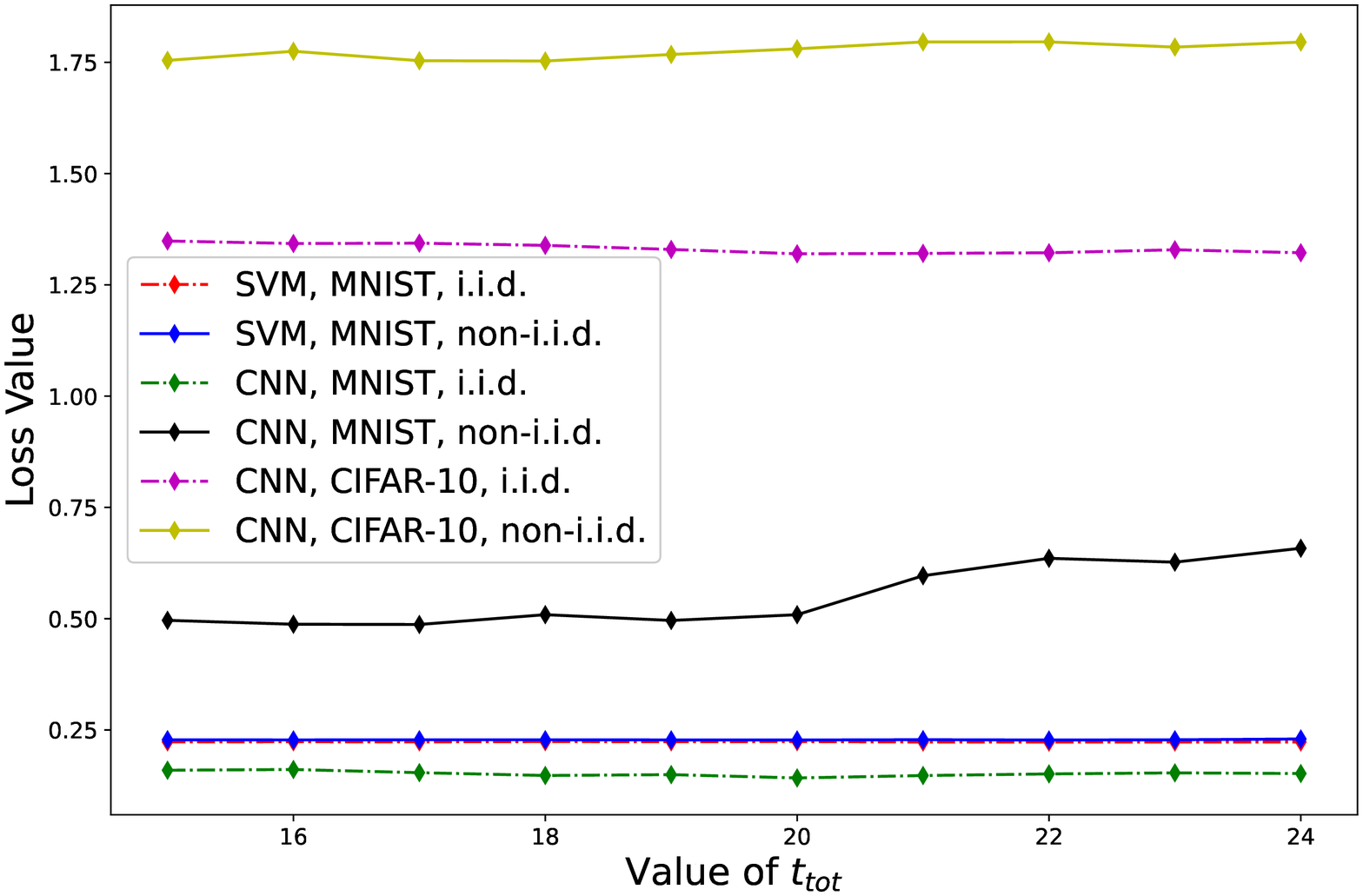}%
\label{fig7_first_case}}\hspace{10mm}
\subfloat[]{\includegraphics[width=3in]{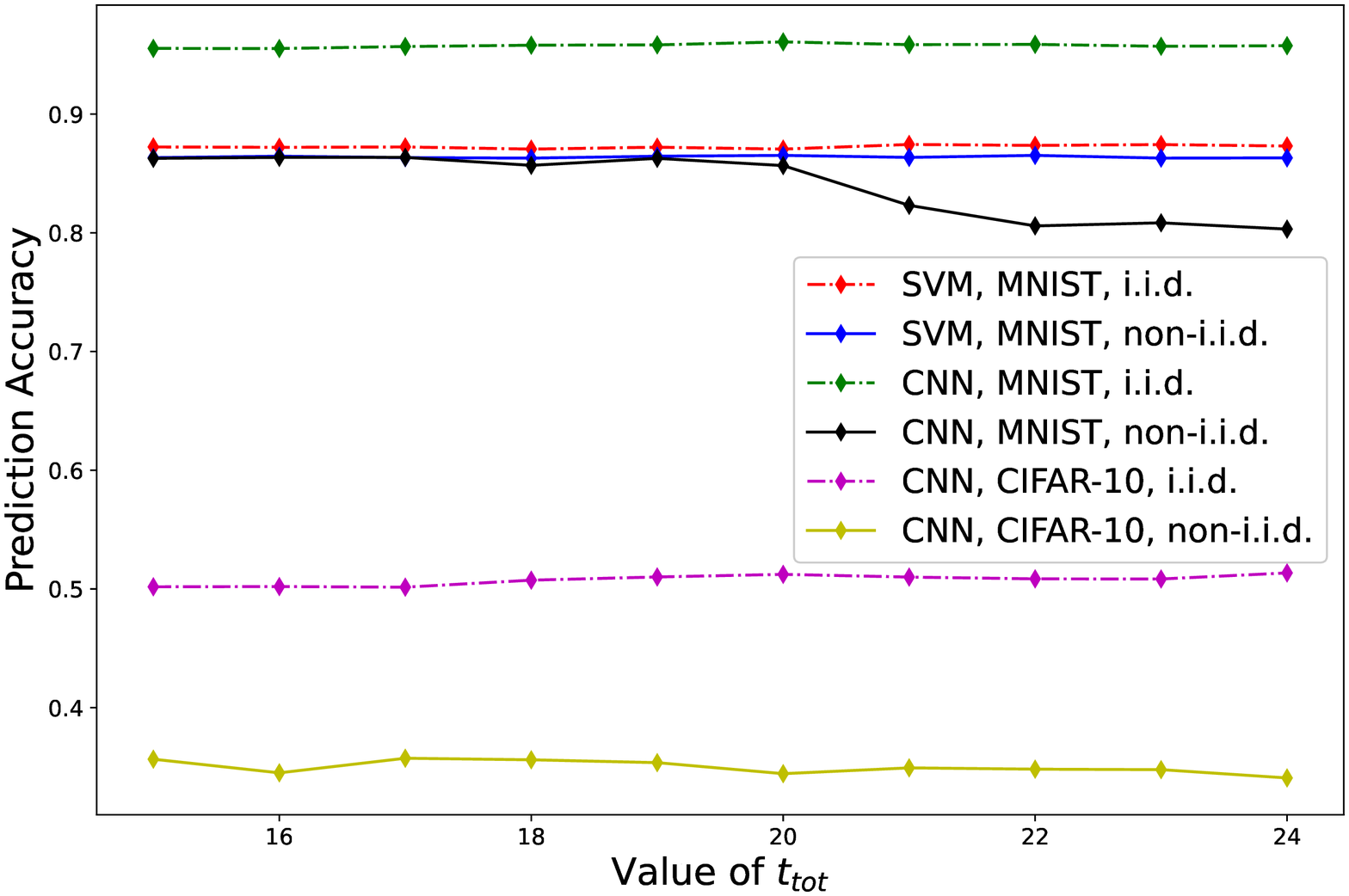}%
\label{fig7_second_case}}
\label{fig7_sim}
\subfloat[]{\includegraphics[width=3in]{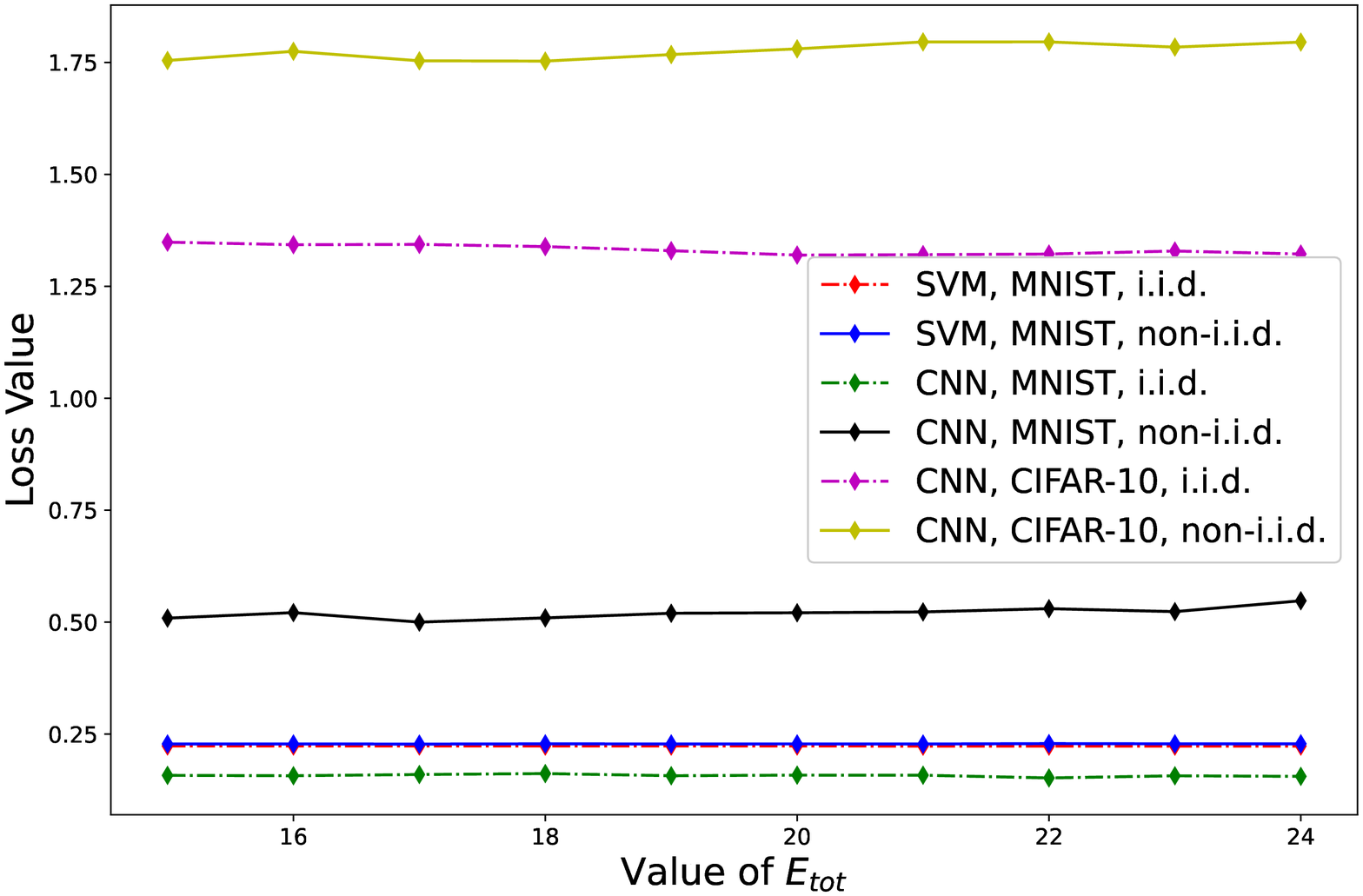}%
\label{fig8_first_case}}\hspace{10mm}
\subfloat[]{\includegraphics[width=3in]{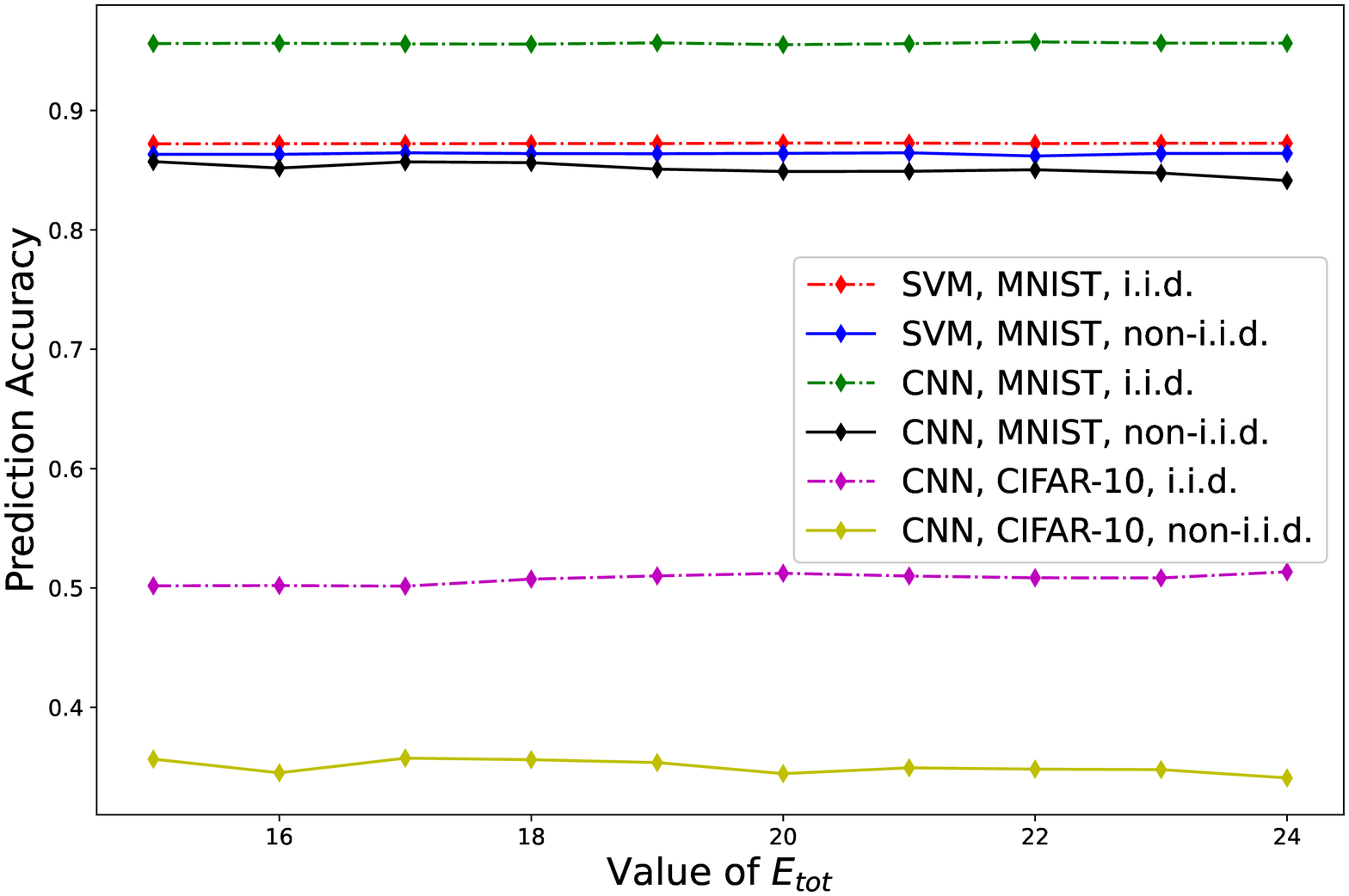}%
\label{fig8_second_case}}
\caption{Loss value and prediction accuracy for different cases. (a) Loss value with fixed $E_{tot}$ and different $t_{tot}$'s. (b) Prediction accuracy with fixed $E_{tot}$ and different $t_{tot}$'s. (c) Loss value with fixed $t_{tot}$ and different $E_{tot}$'s. (d) Prediction accuracy with fixed $t_{tot}$ and different $E_{tot}$'s.}
\label{fig8_sim}
\end{figure*}

\subsection{Results and Observations}
\textit{1) Impact of non-i.i.d. Datasets} 

In the first part of our simulations, both SVM and CNN are trained for every user. We compare the loss value and prediction accuracy with $\tau \in \{1,\cdots,100\}$ in different cases. The results are shown in Fig. 1. 

We see that in different cases with the i.i.d. dataset, as $\tau$ increases, the loss value decreases and predict accuracy improves. In the CNN model, the optimal $\tau$ is large and close to $100$ while close to $20$ in the SVM model. However, we see the loss value of CNN is stable when $\tau$ is more than $20$. Therefore, to avoid overfitting, we set $\tau_{max}$ to be $20$. In the non-i.i.d. dataset, the curves for the loss value represent a ``U'' shape. The optimal $\tau$'s are close to $10$, which is consistent with the analysis in Theorem $2$. It indicates that training model with the non-i.i.d. data need more aggregations (e.g. a lower $\tau$). With the parameters setup in our simulations, the theoretical optimal $\tau$'s in different cases are shown in Table V, which almost coincide with the numerical results in Fig. 1.

\begin{table}[!t]
\caption{$\tau^*$ in different cases\label{tab:table5}}
\centering
\begin{tabular}{cc}
\hline
Case & $\tau^*$ \\
\hline
SVM + i.i.d. + MNIST & 17 \\
SVM + non-i.i.d. + MNIST & 10 \\
CNN + i.i.d. + MNIST & 20 \\
CNN + non-i.i.d. + MNIST & 17 \\
CNN + i.i.d. + CIFAR-10 & 20 \\
CNN + non-i.i.d. + CIFAR-10 & 13 \\
\hline
\end{tabular}
\end{table}

\textit{2) Impact of $t_{tot}$ and $E_{tot}$} 

In the second part of our simulations, we show the performance of different models, datasets, and data distributions with varying $t_{tot}$ and $E_{tot}$. The results are shown in Fig. 2.

Since all cases have already converged, we can evaluate the performance of our proposed update algorithm in these cases. Compared with SVM, the performance of CNN is significantly influenced by the data distribution. Besides, with the i.i.d. dataset, the CNN performs better than the SVM. On the contrary, the SVM outperforms the CNN with the non-i.i.d. dataset. Because CNN is a more complex model (i.e. more parameters) than SVM, it can perform better than SVM ideally. However, CNN is also easy to overfit with non-i.i.d. data, so that CNN performs worse than SVM in this case.

\textit{3) Performance Comparison}

In the third part of our simulations, we compare the proposed algorithm with the baseline from \cite{ref32}, where the optimal $\tau$ is given in Table V. Specifically, the following three categories of comparisons are considered, and their corresponding results are presented in Figs. 3-5.

\begin{itemize}
\item{With the same data distribution and dataset, we compare the performance with the baseline for different models.}
\item{With the same model and dataset, we compare the performance with the baseline for different data distributions.}
\item{With the same data distribution and model, we compare the performance with the baseline for different datasets.}
\end{itemize}

In Fig. 3 and Fig. 4, both the SVM and the CNN are trained with MNIST, which converge faster than the baseline. For instance, in the SVM with $\tau=17$ and i.i.d. MNIST as well as $\tau=10$ and non-i.i.d. MNIST, the loss value converges after $50$ local trainings. However, the number of local trainings in the baseline is around $200$. On the other hand, the prediction accuracy for both the proposed algorithm and the baseline is close to $0.85$ after convergence. In particular, in the CNN with MNIST, the loss value converges after $20$ local trainings with  i.i.d data and $\tau=20$ while  $100$ with non-i.i.d. data and $\tau=17$. However, these numbers in the baseline are close to $20$ and $200$, respectively. Therefore, the convergence speed of our proposed algorithm and the baseline in CNN with MNIST are close to each other. Nevertheless, our proposed algorithm outperforms the baseline in terms of prediction accuracy.

\begin{figure*}[!t]
\centering
\subfloat[]{\includegraphics[width=1.8in]{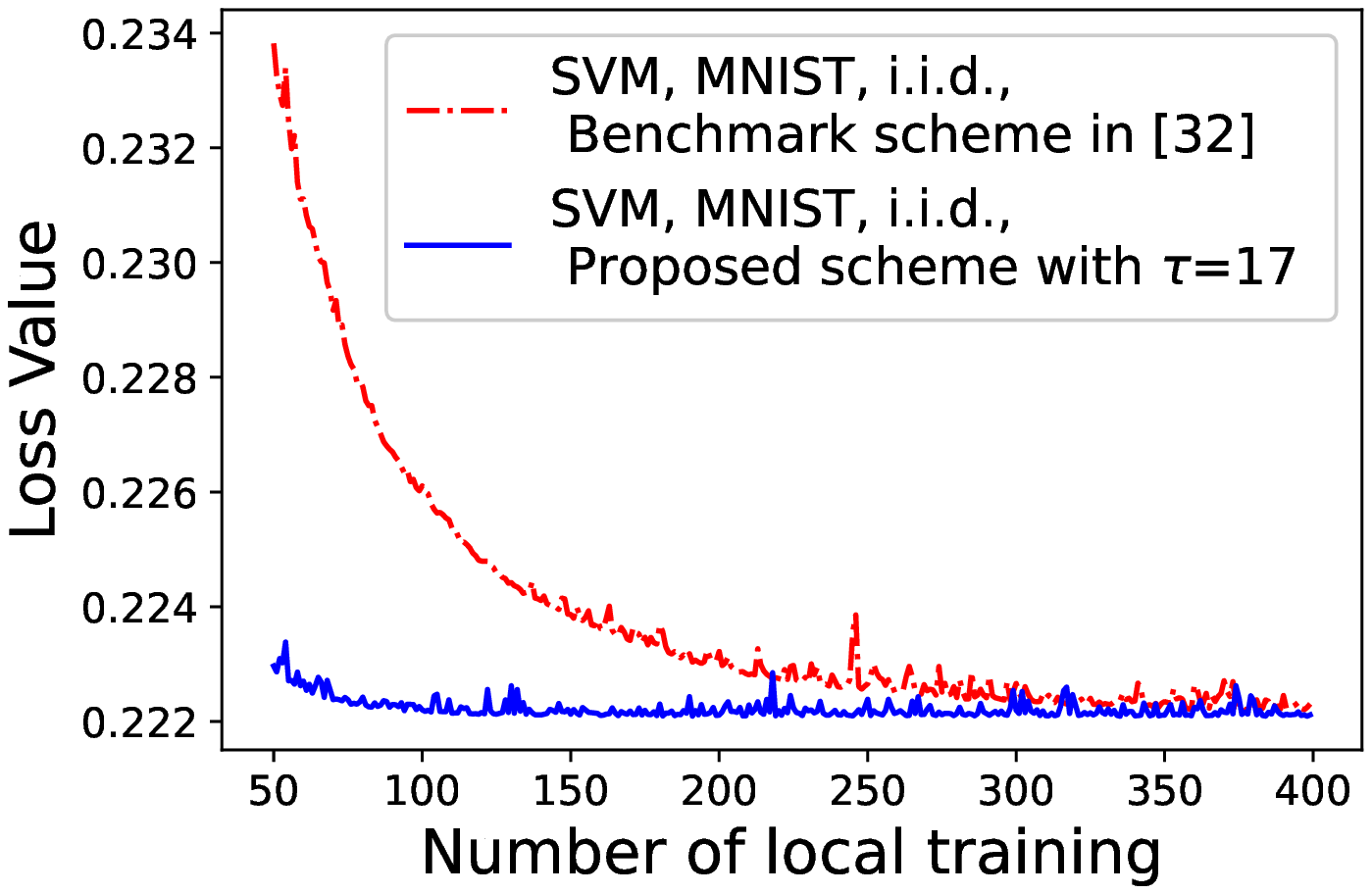}%
\label{fig4_first_case}}
\subfloat[]{\includegraphics[width=1.8in]{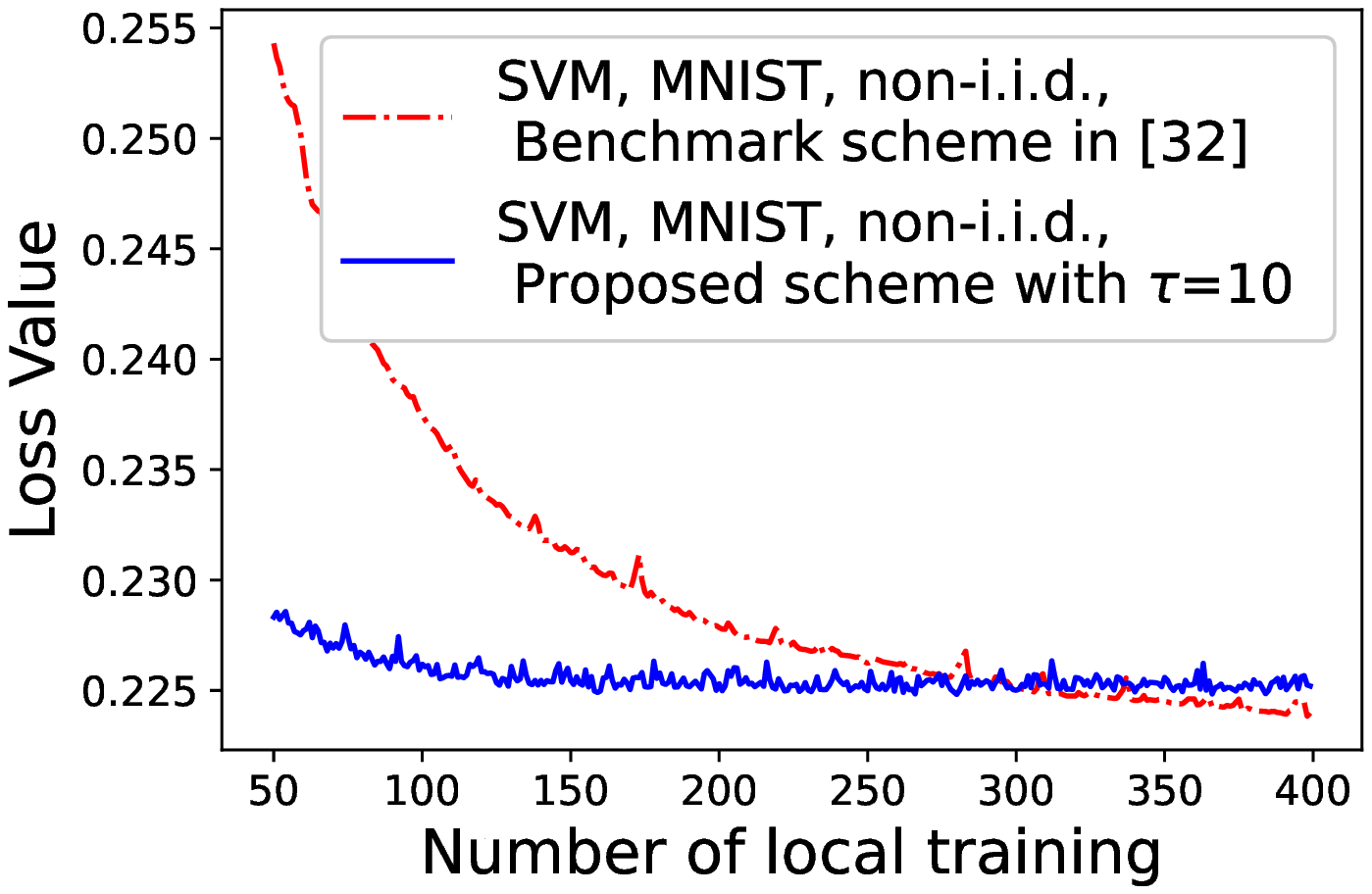}%
\label{fig4_second_case}}
\subfloat[]{\includegraphics[width=1.8in]{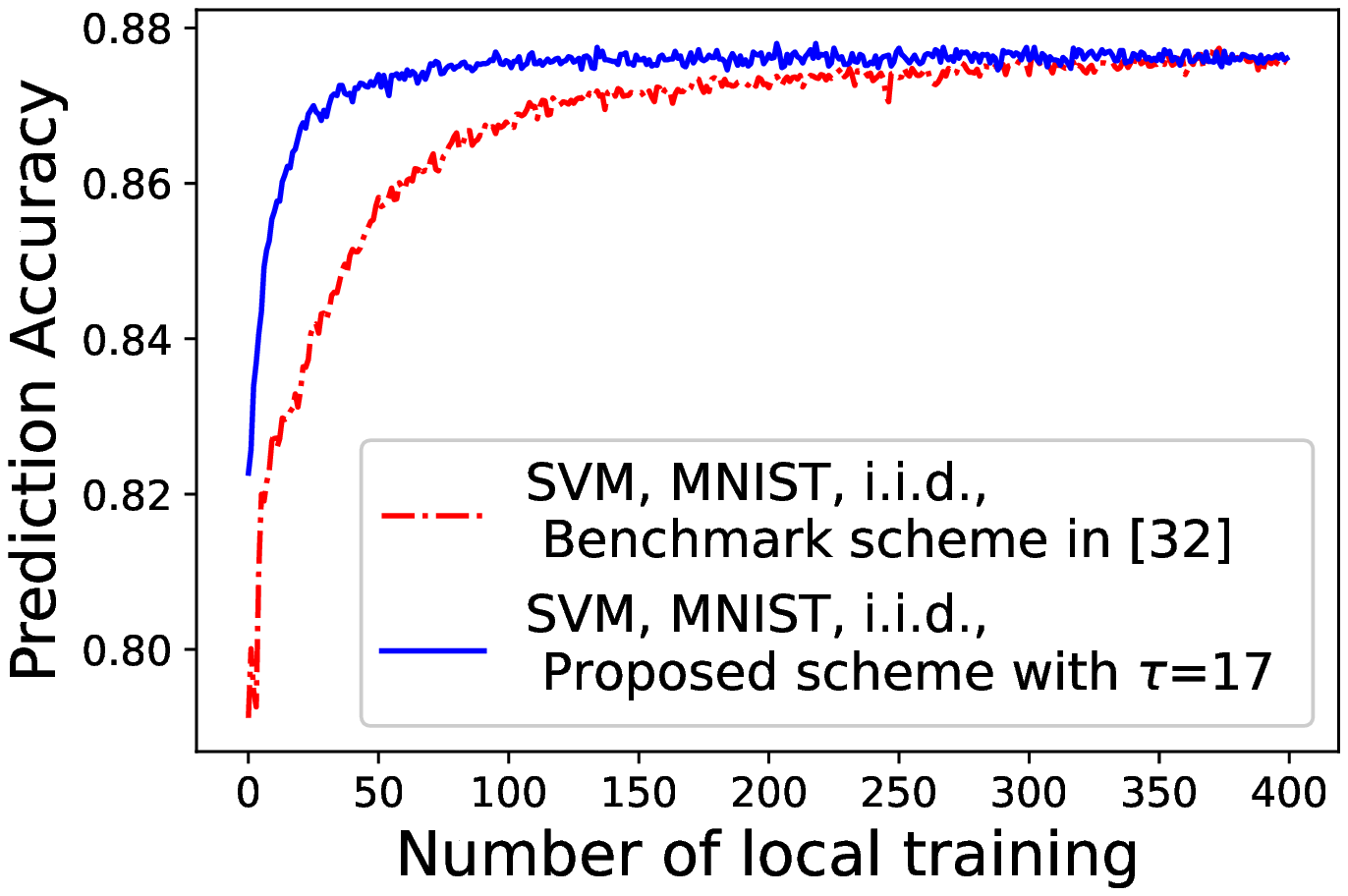}%
\label{fig4_third_case}}
\subfloat[]{\includegraphics[width=1.8in]{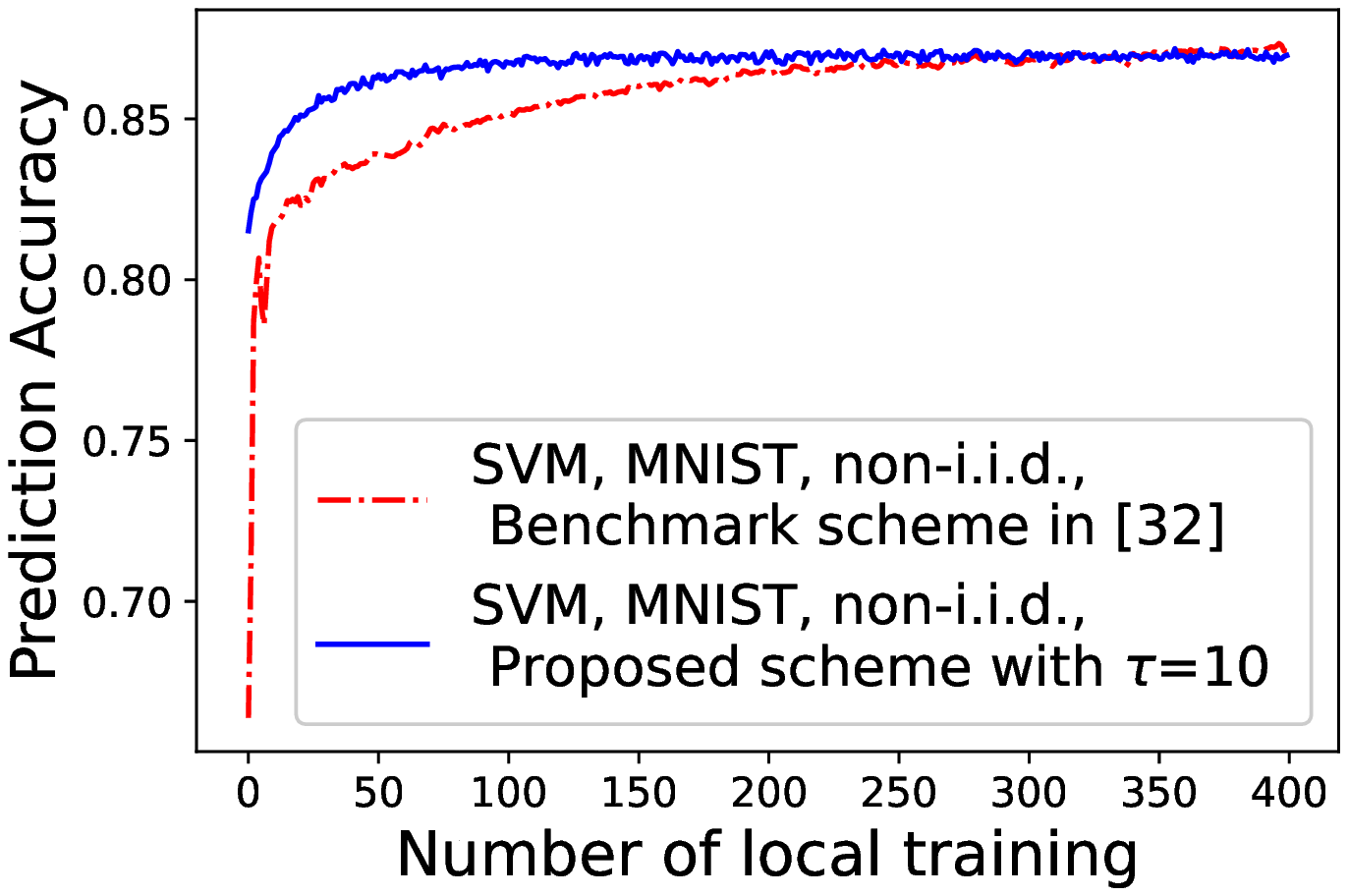}%
\label{fig4_fourth_case}}
\caption{Comparison of loss value and prediction accuracy for SVM with MNIST. 
(a) Loss value with the i.i.d. dataset. (b) Loss value with the non-i.i.d. dataset. 
(c) Prediction accuracy with the i.i.d. dataset.
(d) Prediction accuracy with the non-i.i.d. dataset.}
\label{fig4_sim}
\end{figure*}

\begin{figure*}[!t]
\centering
\subfloat[]{\includegraphics[width=1.8in]{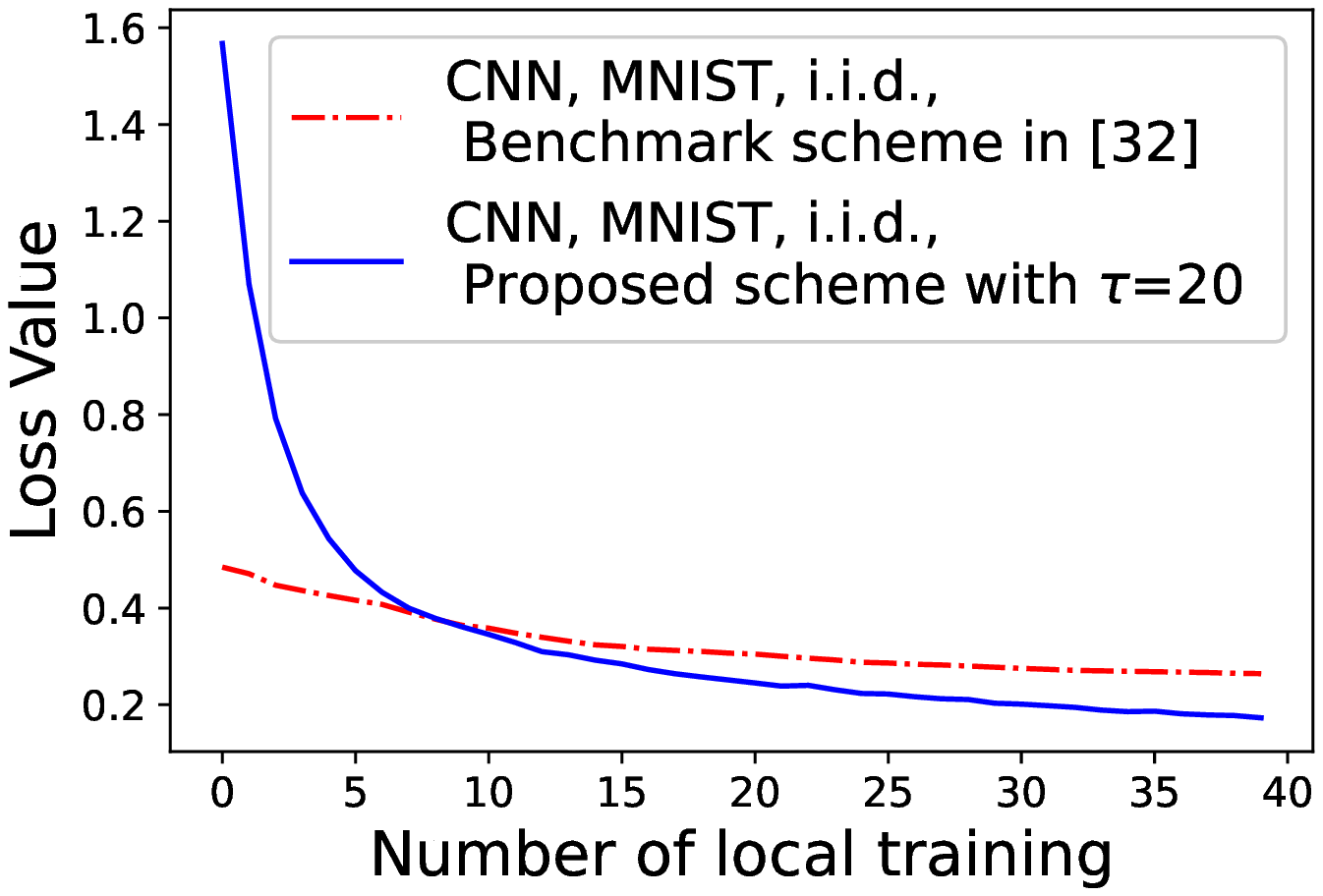}%
\label{fig5_first_case}}
\subfloat[]{\includegraphics[width=1.8in]{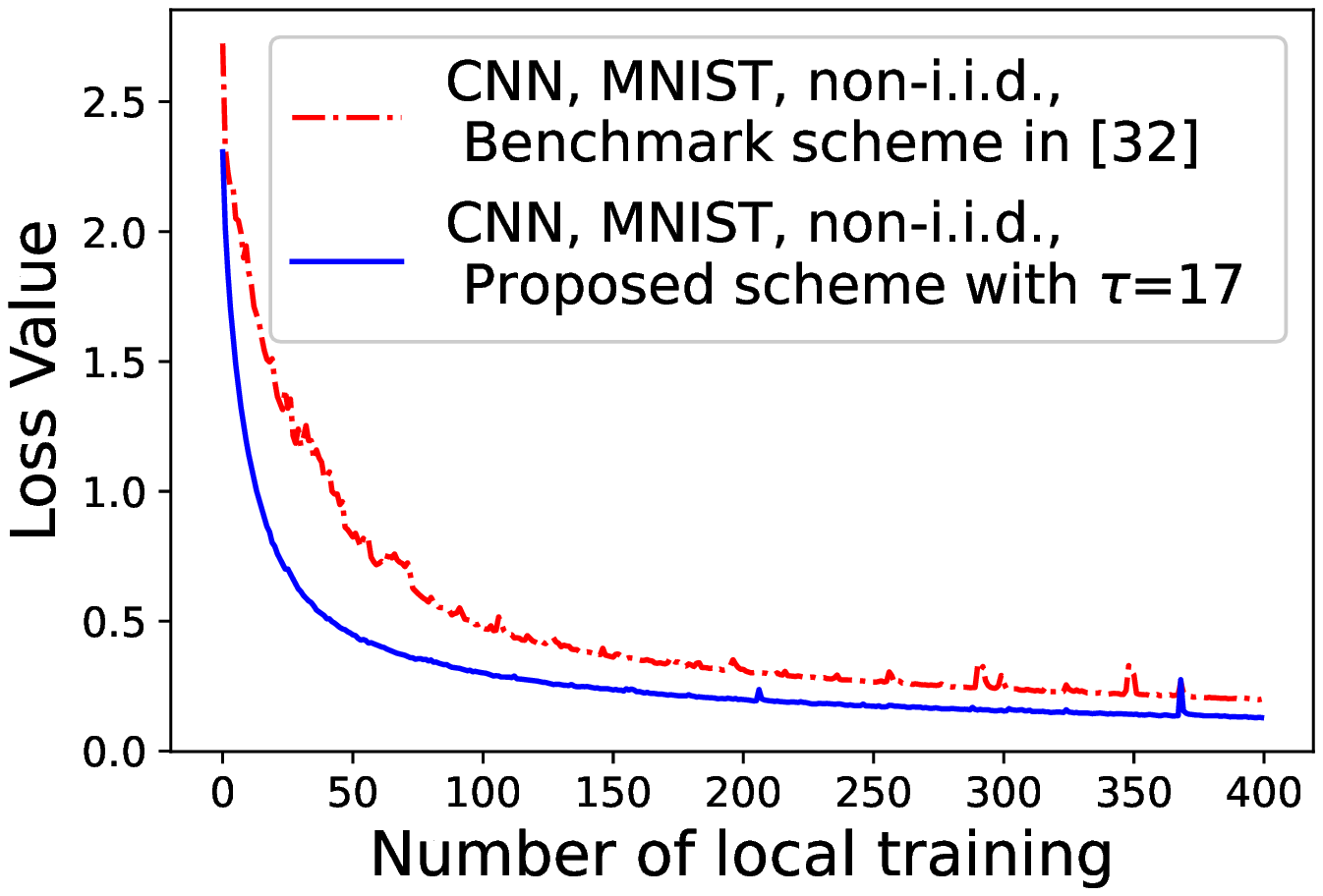}%
\label{fig5_second_case}}
\subfloat[]{\includegraphics[width=1.8in]{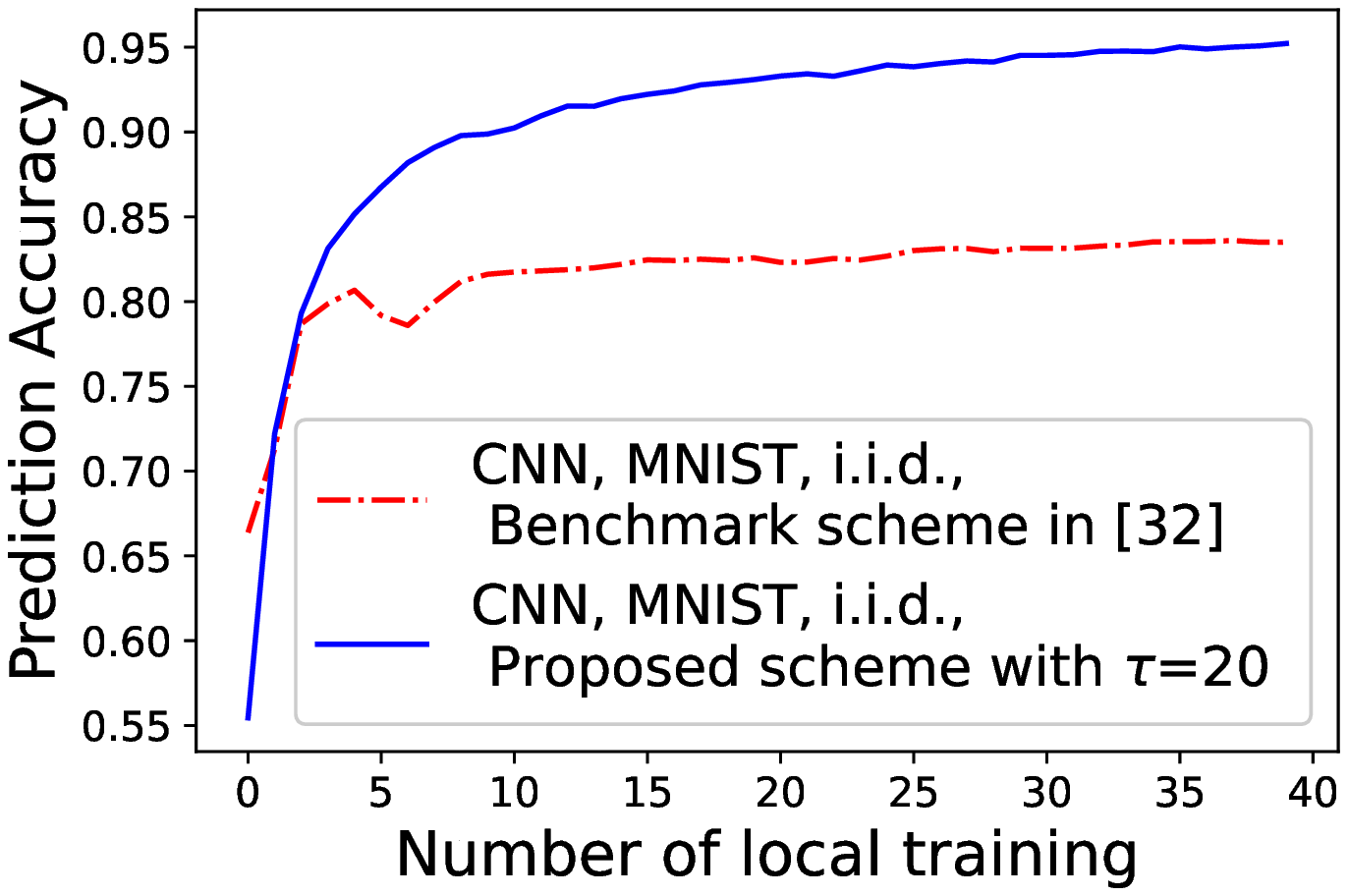}%
\label{fig5_third_case}}
\subfloat[]{\includegraphics[width=1.8in]{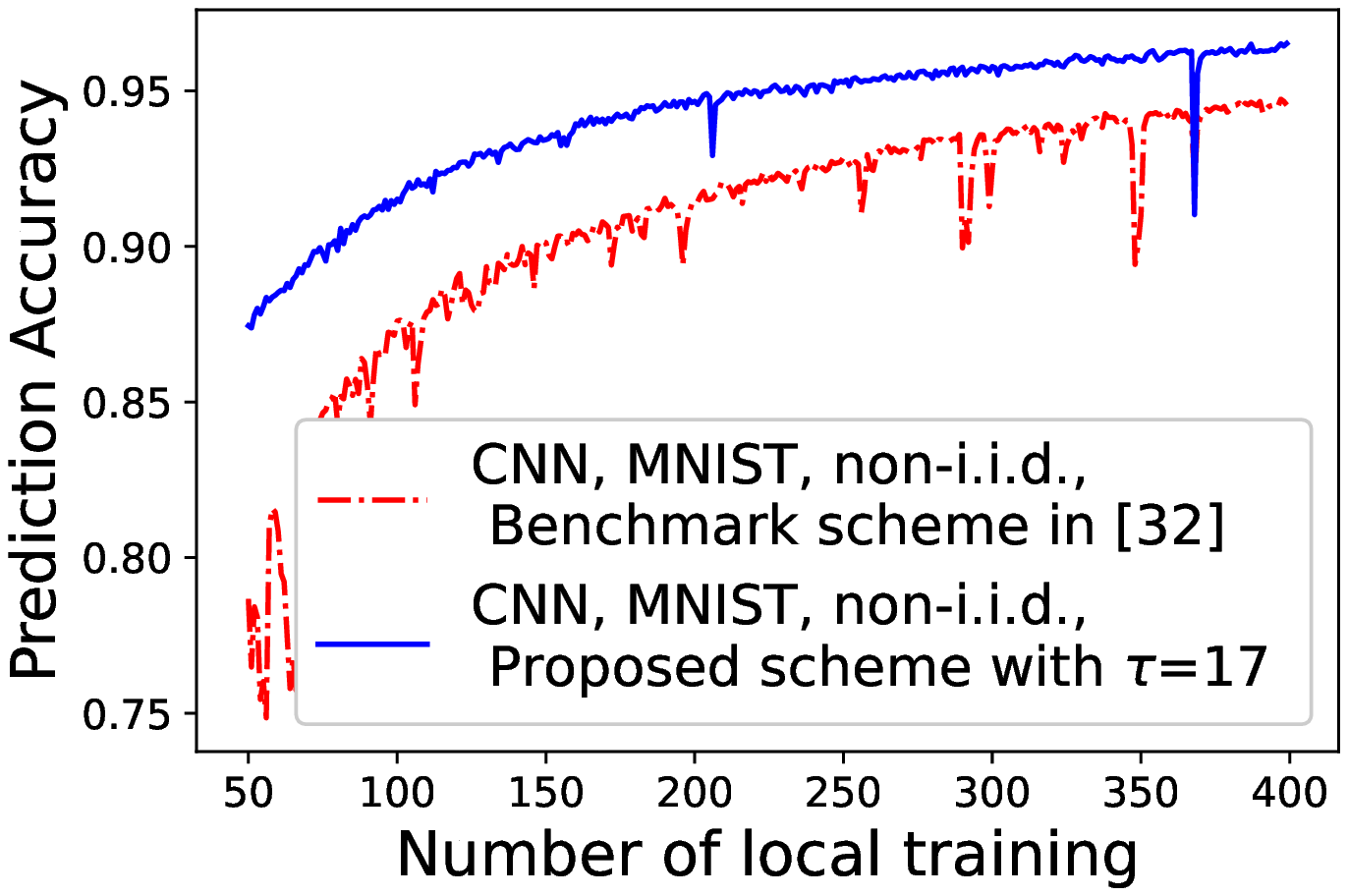}%
\label{fig5_fourth_case}}
\caption{Comparison of loss value and prediction accuracy for CNN with MNIST. 
(a) Loss value with the i.i.d. dataset. (b) Loss value with the non-i.i.d. dataset. 
(c) Prediction accuracy with the i.i.d. dataset.
(d) Prediction accuracy with the non-i.i.d. dataset.}
\label{fig5_sim}
\end{figure*}

\begin{figure*}[!t]
\centering
\subfloat[]{\includegraphics[width=1.8in]{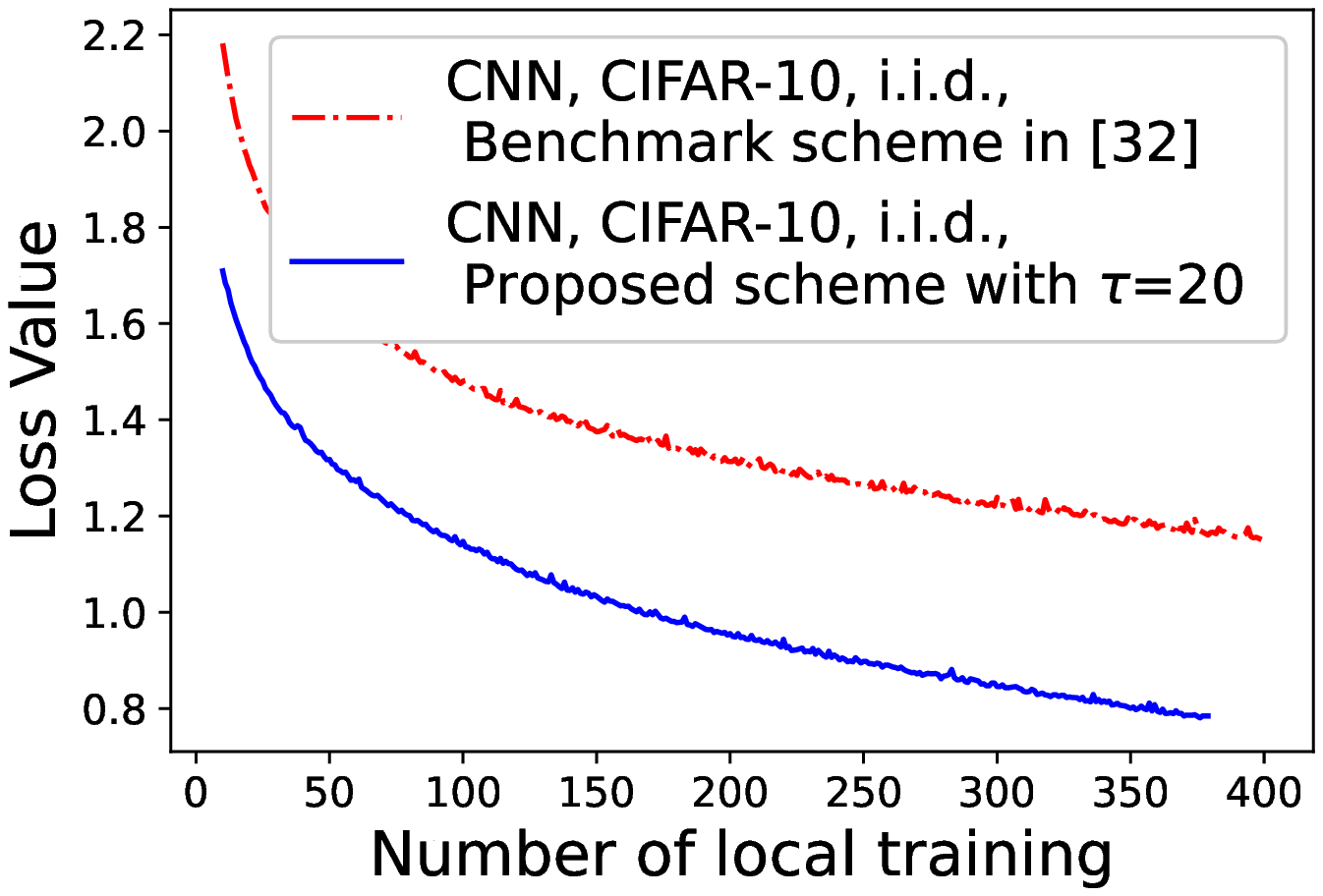}%
\label{fig6_first_case}}
\subfloat[]{\includegraphics[width=1.8in]{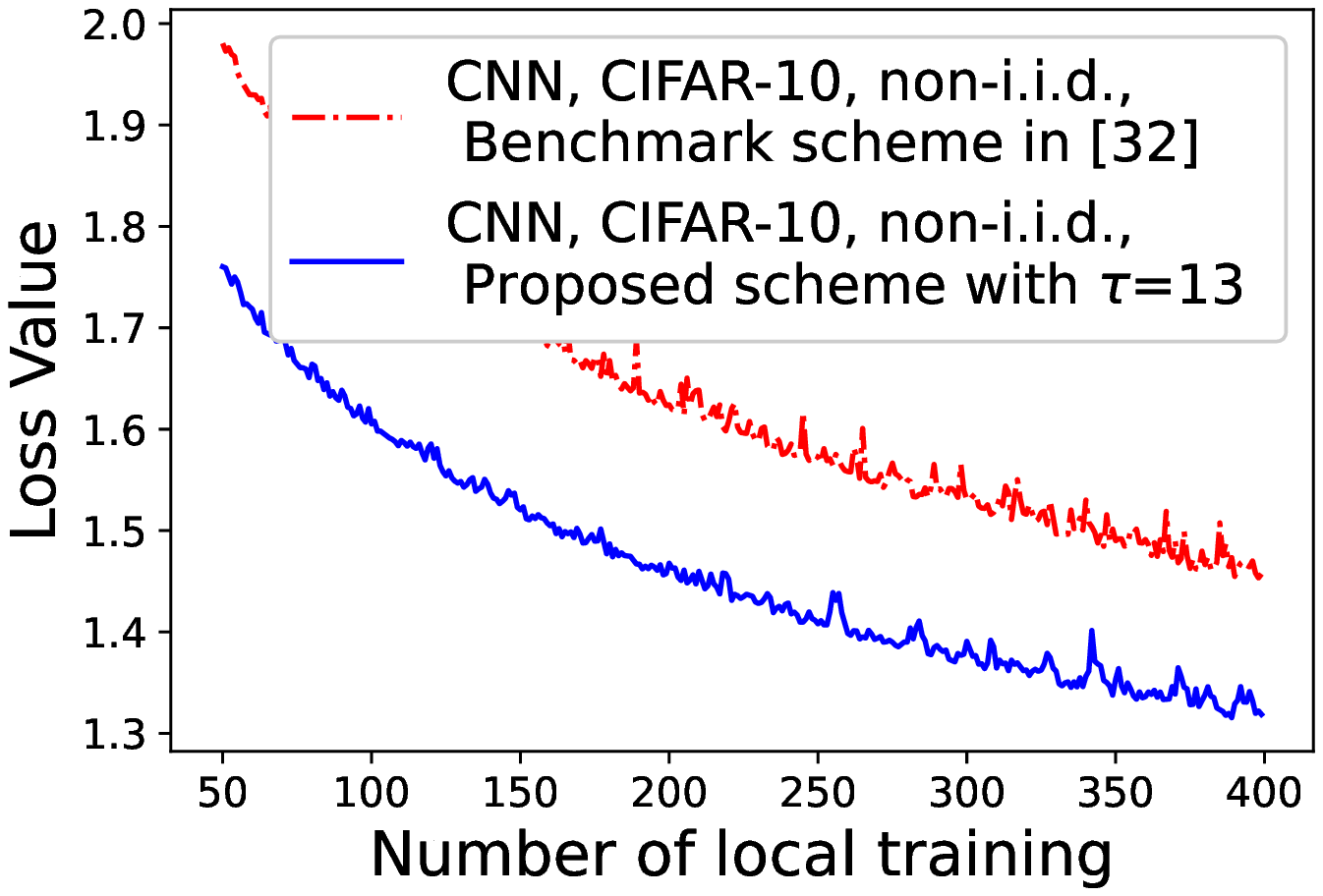}%
\label{fig6_second_case}}
\subfloat[]{\includegraphics[width=1.8in]{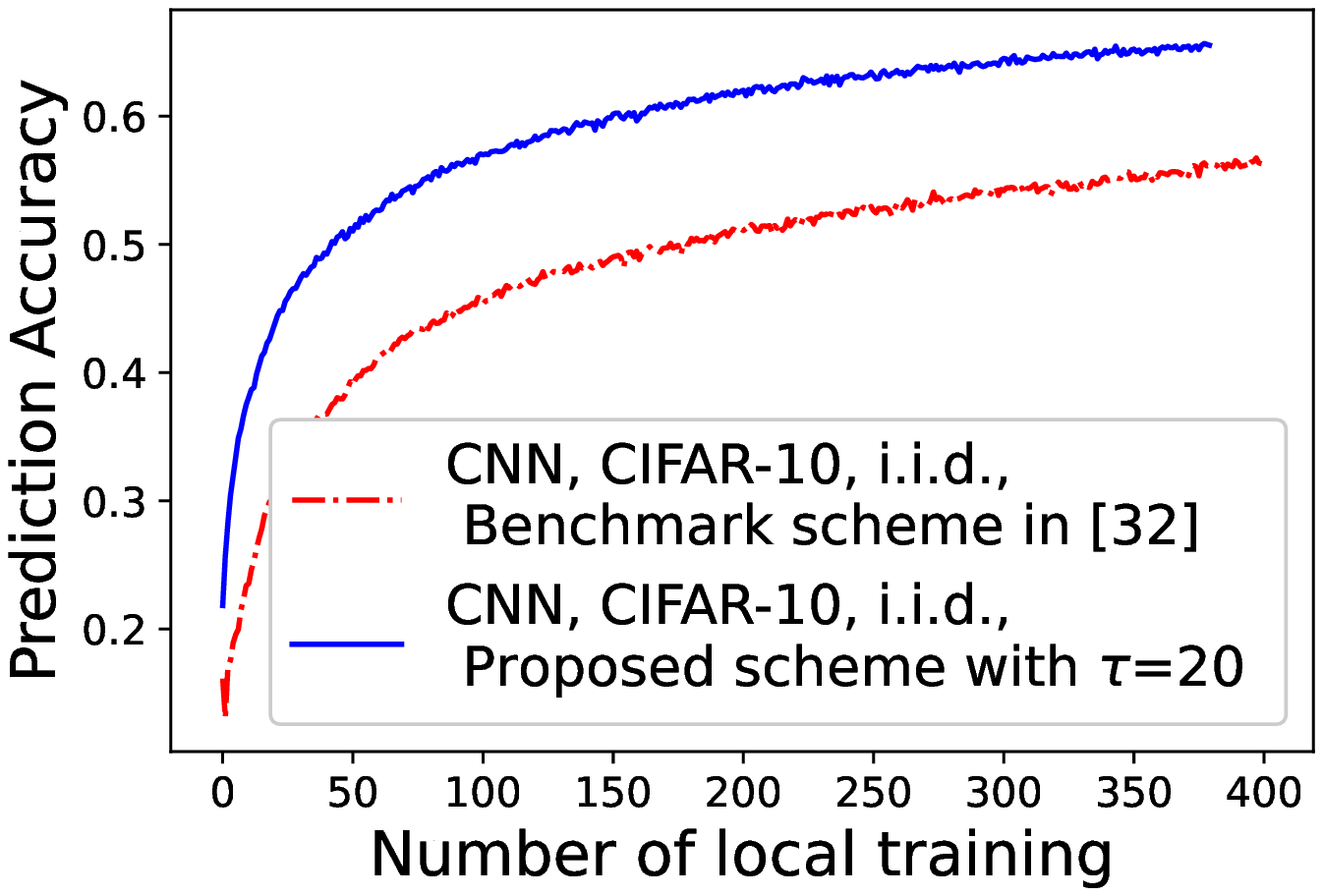}%
\label{fig6_third_case}}
\subfloat[]{\includegraphics[width=1.8in]{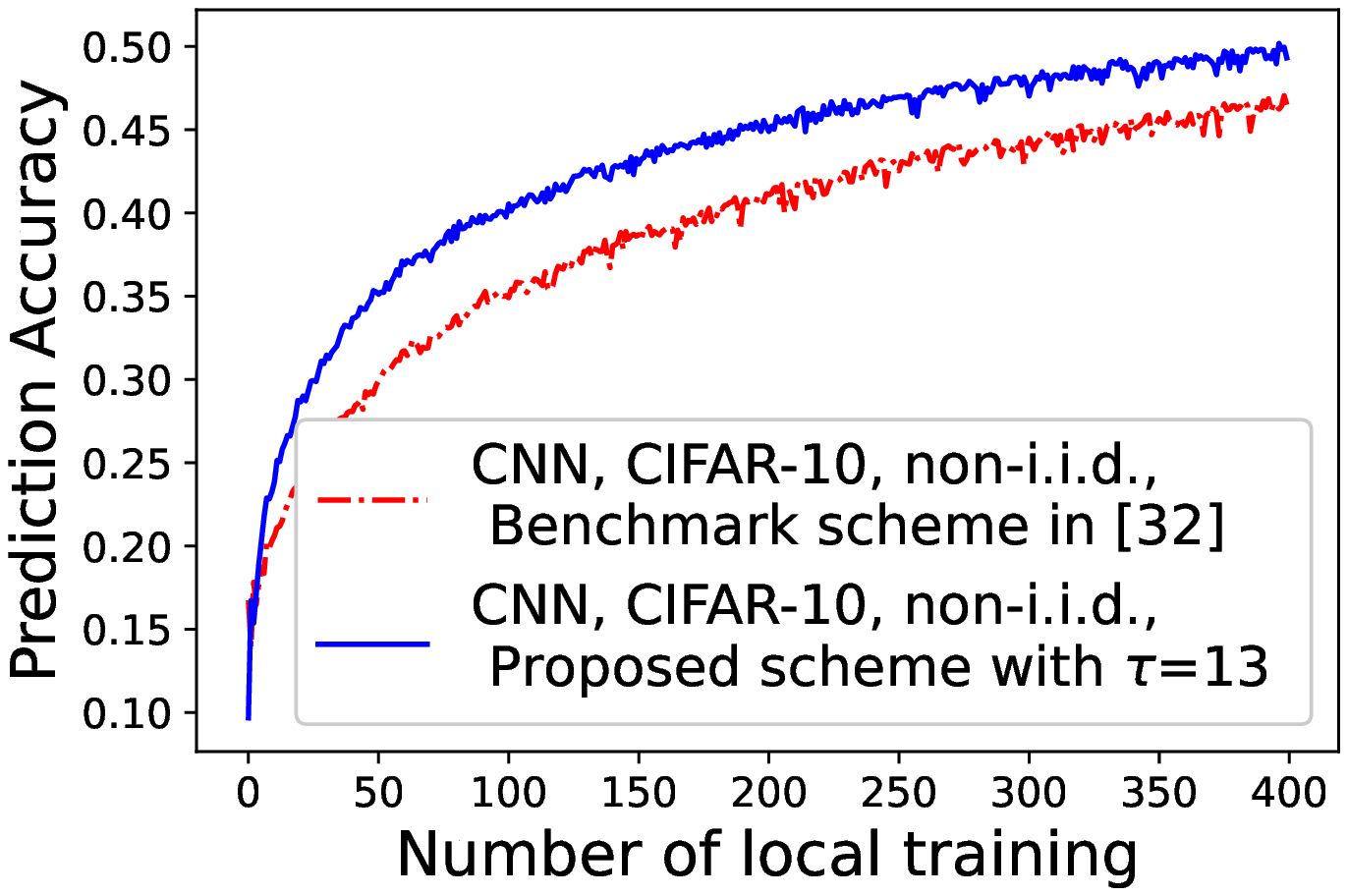}%
\label{fig6_fourth_case}}
\caption{Comparison of loss value and prediction accuracy for CNN with CIFAR-10. 
(a) Loss value with the i.i.d. dataset. (b) Loss value with the non-i.i.d. dataset. 
(c) Prediction accuracy with the i.i.d. dataset.
(d) Prediction accuracy with the non-i.i.d. dataset.}
\label{fig6_sim}
\end{figure*}

In Fig. 3 and Fig. 4, the results not only reflect the difference between SVM and CNN, but also the difference of data distributions. The convergence speed is faster for the i.i.d. data than that for the non-i.i.d data with the SVM and CNN. For the  non-i.i.d. dataset, the proposed algorithm in the SVM and that of the baseline converge after $50$ and $200$ local trainings, respectively. However, these numbers in the non-i.i.d. data are around $100$ and $250$. Furthermore, this phenomenon are also observed in CNN with MNIST. These indicate that the convergence speed for the non-i.i.d. data is slower than that for the i.i.d. data.

In Fig. 4 and Fig. 5, it is seen that both the proposed algorithm and the baseline converge faster with MNIST than CIFAR-10. In the CNN model with MNIST, it converges after $150$ local trainings. However, with CIFAR-10, it needs $200$ local trainings for convergence. These indicate that the convergence speed for the easy task with a simple dataset (i.e., less channels and smaller size per sample) is faster than that for the difficult task with a complex dataset.

\section{Conclusions}
In this paper, we have studied the optimal trade-off between computation and communication in FL with resource restriction. Our goal is to minimize the global loss function while satisfying both the delay and energy consumption requirements. To solve this problem, we have transformed the loss function and derived its closed-form expression in a tractable manner. Through this, we have been able to obtain the optimal number of the local trainings and global aggregations. Furthermore, we have proposed an update algorithm for the implementation of the FL framework. Simulation results have confirmed the effectiveness of the proposed algorithm, i.e., faster convergence and significant performance gain over the conventional ones.

{\appendices

\section{Proof of Theorem 1}

Because $t_{i,tr}-a_i\tau\sim \mathrm{exp}(\frac{\mu}{\tau})$, we know that the probability distribution function of $t_{i,tr}-a_i\tau$ is

\setcounter{equation}{20} 
\begin{equation}
\label{eq20}
{F_1(t)=\mathbb{P}(t_{i,tr}-a_i\tau\leq t)} = \begin{cases}
1-e^{-\frac{\mu}{\tau}t},& t\geq 0, \\ 
{0,}&{\text{otherwise}}, 
\end{cases}
\end{equation}
Thus, when $t\geq 0$, we know that the probability distribution function of $\max\{t_{i,tr}-a_i\tau\}$ is

\begin{equation}
\label{eq21}
\begin{aligned}
F_2(t)&=\mathbb{P}(\max\{t_{i,tr}-a_i\tau\}\leq t) \\
&=\mathbb{P}(t_{i,tr}-a_i\tau\leq u, i=1,2,\dots,N) \\
&=\prod_{i=1}^{N}\mathbb{P}(t_{i,tr}-a_i\tau\leq t) \\
&=[\mathbb{P}(t_{i,tr}-a_i\tau\leq t) ]^{N}\\
&=[1-e^{-\frac{\mu}{\tau}t}]^{N}
\end{aligned}
\end{equation}
It means the probability distribution function of $\max\{t_{i,tr}-a_i\tau\}$ is

\begin{equation}
\label{eq22}
\begin{aligned}
F_2(t)&=\mathbb{P}(\max\{t_{i,tr}-a_i\tau\}\leq t) \\
&= \begin{cases}
[1-e^{-\frac{\mu}{\tau}t}]^{N},&t\geq 0, \\ 
{0,}&{\text{otherwise}}. 
\end{cases}
\end{aligned}
\end{equation}
Therefore, the probability density function of $\max\{t_{i,tr}-a_i\tau\}$ is

\begin{equation}
\label{eq23}
\begin{aligned}
f(t)&=F_2'(t) \\
&= \begin{cases}
\frac{N\mu}{\tau}[1-e^{-\frac{\mu}{\tau}t}]^{N-1}e^{-\frac{\mu}{\tau}t},&t\geq 0, \\ 
{0,}&{\text{otherwise}}. 
\end{cases}
\end{aligned}
\end{equation}
Thus we have the mathematical expectation of $\max\{t_{i,tr}-a_i\tau\}$ is

\begin{equation}
\label{eq24}
\begin{aligned}
&\mathbb{E}(\max\{t_{i,tr}-a_i\tau\})=\int_{0}^{+\infty}tf(t)dt \\
&= \int_{0}^{+\infty}t\cdot N\cdot[1-e^{-\frac{\mu}{\tau}t}]^{N-1}\cdot\frac{\mu}{\tau}e^{-\frac{\mu}{\tau}t} dt.
\end{aligned}
\end{equation}
Then, according to the binomial theorem and the method of integration by parts, we have

\begin{equation}
\label{ETR}
\begin{aligned}
&\int_{0}^{+\infty}t\cdot N\cdot[1-e^{-\frac{\mu}{\tau}t}]^{N-1}\cdot\frac{\mu}{\tau}e^{-\frac{\mu}{\tau}t} dt\\
&=\frac{N\tau}{\mu}\sum_{i=1}^{N}\frac{C_{N-1}^{i-1}(-1)^{i-1}}{i^2}.
\end{aligned}
\end{equation}
Thus, we know that

\begin{equation}
\label{eq25}
\mathbb{E}(\max\{t_{i,tr}\})\leq\frac{N\tau}{\mu}\sum_{i=1}^{N}\frac{C_{N-1}^{i-1}(-1)^{i-1}}{i^2}+a\tau.
\end{equation}

This completes the proof.

\section{Proof of Lemma 2}
By using the triangle inequality, we have

\begin{equation}
\label{eq26}
\vert\Vert\nabla{F_i}(\mathbf{w})\Vert - \Vert\nabla{F}(\mathbf{w})\Vert\vert \leq \Vert\nabla{F_i}(\mathbf{w}) - \nabla{F}(\mathbf{w})\Vert.
\end{equation}

On one hand, by combining \eqref{eq16} and Assumption 1(b), we have
\begin{equation}
\label{eq27}
\big|\Vert\nabla{F_i}(\mathbf{w})\Vert - \Vert\nabla{F}(\mathbf{w})\Vert\big| \leq \delta_{{i}}.
\end{equation}

On the other hand, by combining \eqref{eq17} and Assumption 2, we have
\begin{equation}
\label{eq28}
\Vert\nabla{F_i}(\mathbf{w})\Vert \leq \nabla {F^*} + \delta_{{i}}.
\end{equation}
Thus, we can arrive at \eqref{eq10}. This completes the proof.

\section{Proof of Theorem 2}
We prove Theorem 1 by using the induction. Let $j$ be the index of induction and $j=t-1$, we assume that
\begin{equation}
\label{eq29}
\begin{aligned}
\Vert\Tilde{\mathbf{w}}_{{i}}({t-\mathrm{1}})-&\mathbf{w}({T})\Vert \leq g_i(t-\mathrm{1}).
\end{aligned}
\end{equation}

In the following, we consider the following two cases $t=k\tau$ and $t\neq k\tau$. 

On one hand, when $t=k\tau$, because of \eqref{eq3} and \eqref{eq4}
\begin{equation}
\label{eq30}
\begin{aligned}
\Tilde{\mathbf{w}}_{{i}}({t})&=\frac{\sum\limits_{i=1}^{N}D_i\mathbf{w}_i(t)}{D}\\
&=\frac{\sum\limits_{i=1}^{N}D_i(\mathbf{w}_i(t-1))}{D}-\frac{\sum\limits_{i=1}^{N}D_i(\nabla F_i(\mathbf{w}_i(t-\mathrm{1})))}{D}\\
&=\Tilde{\mathbf{w}}_{{i}}({t-\mathrm{1}})-\nabla F(\Tilde{\mathbf{w}}_i(t-\mathrm{1})).
\end{aligned}
\end{equation}

Therefore, we have
\begin{equation}
\label{eq31}
\begin{aligned}
\Vert\Tilde{\mathbf{w}}_{{i}}({t})-&\mathbf{w}({T})\Vert = \Vert\Tilde{\mathbf{w}}_{{i}}({t-\mathrm{\mathrm{1}}})-\eta\nabla{F}(\Tilde{\mathbf{w}}_{{i}}({t-\mathrm{1}}))-\mathbf{w}({T})\Vert\\ 
&\leq \Vert\Tilde{\mathbf{w}}_{{i}}({t-\mathrm{1}})-\mathbf{w}({T})\Vert + \eta\Vert\nabla{F}(\Tilde{\mathbf{w}}_{{i}}({t-\mathrm{1}}))\Vert.
\end{aligned}
\end{equation}

By combing \eqref{eq30}, \eqref{eq31}, and Lemma 2, we have
\begin{equation}
\label{eq32}
\begin{aligned}
\Vert\Tilde{\mathbf{w}}_{{i}}({t})-&\mathbf{w}({T})\Vert \leq g_i(t-\mathrm{1}) + \eta(\nabla {F^*} + \delta_{{i}}).
\end{aligned}
\end{equation}

On the other hand, when $t\neq k\tau$ holds, based on \eqref{eq2} and Lemma 2, we have 

\begin{equation}
\label{eq33}
\begin{aligned}
\Vert\Tilde{\mathbf{w}}_{{i}}({t})-&\mathbf{w}({T})\Vert = \Vert\Tilde{\mathbf{w}}_{{i}}({t-\mathrm{1}})-\eta\nabla{F_i}(\Tilde{\mathbf{w}}_{{i}}({t-\mathrm{1}}))-\mathbf{w}({T})\Vert\\ 
&\leq \Vert\Tilde{\mathbf{w}}_{{i}}({t-\mathrm{1}})-\mathbf{w}({T})\Vert + \eta\Vert\nabla{F_i}(\Tilde{\mathbf{w}}_{{i}}({t-\mathrm{1}}))\Vert\\
&\leq g_i(t-\mathrm{1}) + \eta\nabla {F^*}.
\end{aligned}
\end{equation}

Let $g_i(t) = (\delta_i + \nabla F^*)\eta t - \frac{\tau}{\rho}$. Therefore, when $j=t$, we have
\begin{equation}
\label{eq34}
\begin{aligned}
g_i(t-1) + \nabla F^* < g_i(t-1) + (\delta_i + \nabla F^*) = g_i(t).
\end{aligned}
\end{equation}
Thus, we can arrive at \eqref{eq11}. This completes the proof.

\section{Proof of Theorem 3}
Based on \eqref{eq3} and \eqref{eq4}, we have

\begin{equation}
\label{eq35}
\begin{aligned}
\mathbf{w}({p})&=\frac{\sum\limits_{i=1}^{N}D_i\mathbf{w}_i(p)}{D}\\
&=\frac{\sum\limits_{i=1}^{N}D_i(\mathbf{w}_i(p-\mathrm{1})}{D}-\frac{\sum\limits_{i=1}^{N}D_i(\eta\nabla{F_i}(\mathbf{w}({p-\mathrm{1}})))}{D}\\
&=\mathbf{w}({p-\mathrm{1}})-\eta\nabla{F}(\mathbf{w}({p-\mathrm{1}}))
\end{aligned}
\end{equation}

and

\begin{equation}
\label{eq36}
\begin{aligned}
\Vert\mathbf{w}({p})&-\mathbf{w}({T})\Vert\\
&=\Vert\mathbf{w}({p-\mathrm{1}})-\eta\nabla{F}(\mathbf{w}({p-\mathrm{1}}))-\mathbf{w}({T})\Vert\\
&\leq \Vert\mathbf{w}({p-\mathrm{1}})-\mathbf{w}({T})\Vert + \eta\Vert\nabla{F}(\mathbf{w}({p-\mathrm{1}}))\Vert.
\end{aligned}
\end{equation}
By using Assumption 2, \eqref{eq36} becomes
\begin{equation}
\label{eq37}
\begin{aligned}
\Vert\mathbf{w}({p})-\mathbf{w}({T})\Vert - \Vert\mathbf{w}({p-\mathrm{1}})-\mathbf{w}({T})\Vert \leq \eta\nabla{F^*},\\
p={2,3,\cdots,t}.
\end{aligned}
\end{equation}
By summing all items of \eqref{eq37} on both sides, the item with $p$ can be deleted and we further have
\begin{equation}
\label{eq38}
\begin{aligned}
 \Vert\mathbf{w}({t})-\mathbf{w}({T})\Vert \leq \Vert\mathbf{w}(\mathrm{1})-\mathbf{w}({T})\Vert + \eta t\nabla{F^*}.\\
\end{aligned}
\end{equation}
Besides, using \eqref{eq3} and Theorem 1, we have
\begin{equation}
\label{eq39}
\begin{aligned}
\Vert\mathbf{w}(\mathrm{1})-&\mathbf{w}({T})\Vert = \Vert\frac{\sum\limits_{i=\mathrm{1}}^{N}{D_i}\mathbf{w}_i(\mathrm{1})}{D}-\mathbf{w}({T})\Vert\\
&\leq \frac{\sum\limits_{i=1}^{N}{D_i}\Vert\mathbf{w}_i(\mathrm{1})-\mathbf{w}({T})\Vert}{D}\\
&\leq \frac{\sum\limits_{i=1}^{N}{D_i}g_i(\mathrm{1})}{D}\\
&= (\delta + \nabla F^*)\eta - \frac{\tau}{\rho}.
\end{aligned}
\end{equation}

By combining \eqref{eq38} and \eqref{eq39}, we have
\begin{equation}
\label{eq40}
\begin{aligned}
\Vert\mathbf{w}({t})-&\mathbf{w}({T})\Vert \leq (\delta + \nabla F^*)\eta - \frac{\tau}{\rho} + \eta\nabla{F^*}t.
\end{aligned}
\end{equation}

Furthermore, by using Lemma 1 and \eqref{eq30}, we have
\begin{equation}
\label{eq41}
\begin{aligned}
{F}(\mathbf{w}({t}))-{F}(\mathbf{w}({T})) \leq \rho\Vert\mathbf{w}({t})-&\mathbf{w}({T})\Vert \\
=\rho\eta(\delta + \nabla F^*) - \tau +  \rho\eta\nabla{F^*}T.
\end{aligned}
\end{equation}

Let $\theta(t)\triangleq{F}(\mathbf{w}({t}))-{F}(\mathbf{w}^*)$, according to Assumption 1 5), we have $({F}(\mathbf{w}({T}))-{F}(\mathbf{w}^*)){\theta(t)}\geq\epsilon^{\textrm{2}}$, or equivalently

\begin{equation}
\label{eq42}
\begin{aligned} -\frac{1}{({F}(\mathbf{w}({T}))-{F}(\mathbf{w}^*))\theta({t})}\geq-\frac{1}{\epsilon^2}.
\end{aligned}
\end{equation}
By combining \eqref{eq31} and \eqref{eq32}, we have

\begin{equation}
\label{eq43}
\begin{aligned}
&\frac{1}{\theta(t)}-\frac{1}{{F}(\mathbf{w}({T}))-{F}(\mathbf{w^*})}\\
&=\frac{{F}(\mathbf{w}({T}))-{F}(\mathbf{w}({t}))}{\theta(t)({F}\mathbf{w}({T}))-{F}(\mathbf{w^*}))}\\
&\geq -\frac{\rho\eta(\delta + \nabla F^*) - \tau +  \rho\eta\nabla{F^*}T}{\epsilon^2},\\
\end{aligned}
\end{equation}
or equivalently
\begin{equation}
\label{eq44}
\begin{aligned}
&\theta(t)\\
&\leq\frac{\epsilon^2({F}(\mathbf{w}({T}))-{F}(\mathbf{w}^*))}{\epsilon^2 - ({F}(\mathbf{w}({T}))-{F}(\mathbf{w}^*))(\rho\eta(\delta + \nabla {F^*}) - \tau + \rho\eta\nabla{F^*}T)}.\\
\end{aligned}
\end{equation}
Since

\begin{equation}
\label{eq45}
\begin{aligned}
\theta(t)&={F}(\mathbf{w}({t}))-{F}(\mathbf{w^*})\\
&\geq {F}(\mathbf{w}({T}))-{F}(\mathbf{w^*}).
\end{aligned}
\end{equation}
we have
\begin{equation}
\label{eq46}
\begin{aligned}
&{F}(\mathbf{w}({T}))-{F}(\mathbf{w^*})\\
&\leq\frac{\epsilon^2({F}(\mathbf{w}({T}))-{F}(\mathbf{w^*}))}{\epsilon^2 - ({F}(\mathbf{w}({T}))-{F}(\mathbf{w^*}))(\rho\eta(\delta + \nabla {F^*}) - \tau + \rho\eta\nabla{F^*}T)}.
\end{aligned}
\end{equation}
However, \eqref{eq46} cannot be directly solved to obtain the upper bound of ${F}(\mathbf{w}({T}))-{F}(\mathbf{w^*})$. To circumvent this problem, we introduce $\epsilon$ on the left hand of \eqref{eq47} and rewrite \eqref{eq47} as

\begin{equation}
\label{eq47}
\begin{aligned}
&{F}(\mathbf{w}({T}))-{F}(\mathbf{w^*})-\epsilon\\
&\leq\frac{\epsilon^2({F}(\mathbf{w}({T}))-{F}(\mathbf{w^*}))}{\epsilon^2 - ({F}(\mathbf{w}({T}))-{F}(\mathbf{w^*}))(\rho\eta(\delta + \nabla {F^*}) - \tau + \rho\eta\nabla{F^*}T)}.
\end{aligned}
\end{equation}
For simplicity, we re-express the constants and variables in \eqref{eq47} as follows:

\begin{equation*}
\begin{cases}
{F}(\mathbf{w}({T}))-{F}(\mathbf{w^*})\triangleq{x},\\ 
\epsilon\triangleq a,\\
\rho\eta(\delta + \nabla {F^*}) - \tau + \rho\eta\nabla{F^*}T\triangleq b.
\end{cases}
\end{equation*}
Therefore, \eqref{eq47} can be written as

\begin{equation}
\label{eq48}
\begin{aligned}
&x-a\leq \frac{a^2x}{a^2-bx},
\end{aligned}
\end{equation}
or equivalently
\begin{equation}
\label{eq49}
\begin{aligned}
\frac{bx^2-abx+a^3}{a^2-bx}\geq 0.
\end{aligned}
\end{equation}
If $a^2-bx<0$, it can be deduced from \eqref{eq38} that $0\leq x-a\leq \frac{a^2x}{a^2-bx}<0$, which leads to a contradiction. Therefore \eqref{eq39} can be written as

\begin{equation}
\label{eq50}
\begin{cases}
bx^2-abx+a^3 \geq 0,\\
a^2-bx > 0.
\end{cases}
\end{equation}

Let $\Delta=a^2(b^2-4ab)$. Then we solve this set of inequalities.

\textbf{Case 1: $\Delta\leq 0$.}

Because $b>0$, we know that $bx^2-abx+a^3 \geq 0$ always holds. Therefore solving \eqref{eq40} is equivalent to solving $a^2-bx> 0$.

Thus in Case 1, we have
\begin{equation}
\label{eq51}
\begin{aligned}
x<\frac{a^2}{b}.
\end{aligned}
\end{equation}

\textbf{Case 2: $\Delta>0$.}

When $\Delta>0$, by solving $bx^2-abx+a^3 \geq 0$, we can obtain $x\geq x_1$ or $x\leq x_2$, where $x_1=\frac{ab+ a\sqrt{(b^2-4ab)}}{2b}$, $x_2=\frac{ab- a\sqrt{(b^2-4ab)}}{2b}$. Therefore, \eqref{eq40} can be written as

\begin{equation}
\begin{cases}
\label{eq52}
x\geq x_1 \quad or\quad x\leq x_2,\\
a^2-bx > 0.
\end{cases}
\end{equation}

Because
\begin{equation}
\label{eq53}
\begin{aligned}
x_2 - \frac{a^2}{b}&=\frac{a}{2b}(b-2a-\sqrt{b^2-4ab})\\
&\geq \frac{a}{2b}(b-2a-\sqrt{b^2-4ab+4a^2})\\
&\geq 0.
\end{aligned}
\end{equation}

Then, \eqref{eq52} reduces to

\begin{equation}
\label{eq54}
\begin{aligned}
x<\frac{a^2}{b}.
\end{aligned}
\end{equation}

By combining these two cases, we have

\begin{equation}
\label{eq55}
\begin{aligned}
\mathit{F}(\mathbf{w}(\mathit{T}))-\mathit{F}(\mathbf{w^*})\leq\frac{\epsilon^2}{\rho\eta(\delta + \nabla {F^*}) - \tau + \rho\eta\nabla{F^*}T}.
\end{aligned}
\end{equation}

This completes the proof.

\section{Proof of Theorem 4}

The Lagrangian function of \eqref{eq14} is 

\begin{equation}
\label{eq60}
\begin{aligned}
L(\tau,K,\lambda_1,\lambda_2)&=\tau-\rho\eta\nabla{F^*}K\tau\\
&+\lambda_1(t_{cm}K+(I_{0}N\mu+a)K\tau^2 - t_{tot})\\
&+\lambda_2(P_{cm}t_{cm}K +E{tr}K\tau - E_{tot}).
\end{aligned}
\end{equation}

Furthermore, the KKT conditions of \eqref{eq60} are 

\begin{equation}
\label{eq61}
\begin{cases}
\frac{\partial L}{\partial \tau}=0,\\
\frac{\partial L}{\partial K}=0,\\
P_{cm}t_{cm}K +E{tr}K\tau - E_{tot}\leq 0,\\
t_{cm}K+(I_{0}N\mu+a)K\tau^2 - t_{tot}\leq 0,\\
\lambda_1(t_{cm}K+(I_{0}N\mu+a)K\tau^2 - t_{tot})=0,\\
\lambda_2(P_{cm}t_{cm}K +E{tr}K\tau - E_{tot})=0,\\
\lambda_1,\lambda_2\geq 0.\\
\end{cases}
\end{equation}

By taking the derivative of \eqref{eq60} with respect to $\tau$ and $K$ respectively, and setting them to be equal to zero, we have

\begin{equation*}
\begin{cases}
\frac{\partial L}{\partial \tau}=1-\rho\eta\nabla{F^*}K+2\lambda_\textrm{1}(I_{0}N\mu+a)K\tau+\lambda_\textrm{2}E_{tr}K=\textrm{0},\\
\frac{\partial L}{\partial K}=-\rho\eta\nabla{F^*}\tau+\lambda_\textrm{1}t_{cm}+\lambda_\textrm{1}(I_{0}N\mu+a)\tau^{2}+\lambda_\textrm{2}P_{cm}t_{cm}\\
\quad\quad\quad+\lambda_\textrm{2}E_{tr}\tau=0.\\
\end{cases}
\end{equation*}

Therefore

\begin{equation}
\label{eq62}
\begin{cases}
\lambda_\textrm{1}(I_{0}N\mu+a)(\tau^{*})^{2}-(\lambda_\textrm{2}E_{tr}-\rho\eta\nabla{F^*})\tau^*\\
\quad\quad\quad+\lambda_\textrm{1}t_{cm}+\lambda_\textrm{2}P_{cm}t_{cm}=0,\\
K^*=\frac{1}{\rho\eta\nabla{F^*}-2\lambda_\textrm{1}(I_{0}N\mu+a)\tau-\lambda_\textrm{2}E_{tr}}.\\
\end{cases}
\end{equation}

Next, we discuss the following four cases regarding \eqref{eq61}.

\textbf{Case 1: }$\lambda_1, \lambda_2=0$

When $\lambda_1, \lambda_2=0$, $\tau^*=0$, so this case is not true.

\textbf{Case 2: }$\lambda_1, \lambda_2\neq0$

When $\lambda_1, \lambda_2\neq0$, the \eqref{eq61} means that

\begin{equation}
\label{eq63}
\begin{cases}
P_{cm}t_{cm}K +E{tr}K\tau - E_{tot}=0,\\
t_{cm}K+(I_{0}N\mu+a)K\tau^2 - t_{tot}=0.\\
\end{cases}
\end{equation}
or equivalently

\begin{equation}
\label{eq64}
\frac{t_{cm}+(I_{0}N\mu+a)\tau^2}{P_{cm}t_{cm} +E_{tr}\tau}=\frac{t_{tot}}{E_{tot}}
\end{equation}
Then we solve \eqref{eq64} and get

\begin{equation}
\label{eq65}
\begin{aligned}
\tau_{1,2}^{*}=\frac{E_{tr}t_{tot}\pm\sqrt{I_{1}}}{2(I_{0}N\mu+a)E_{tot}},
\end{aligned}
\end{equation}
where $I_{1}=E^{2}_{tr}t^{2}_{tot}-4(I_{0}N\mu+a)E_{tot}(t_{cm}E_{tot}-P_{cm}t_{cm}t_{tot})$, and $K^*=\frac{E_{tot}}{P_{cm}t_{cm}+E_{tr}\tau^*}$.

Because minimizing $\tau-\rho\eta\nabla{F^*}K\tau$ requires a smaller $\tau^*$ and a larger $K^*$, we have

\begin{equation}
\label{eq66}
\begin{cases}
\tau_1^*=\frac{E_{tr}t_{tot}-\sqrt{I_{1}}}{2(I_{0}N\mu+a)E_{tot}},\\
K_1^*=\frac{2(I_{0}N\mu+a)E^{2}_{tot}}{2(I_{0}N\mu+a)P_{cm}t_{cm}E_{tot}+E^{2}_{tr}t_{tot}-E_{tr}\sqrt{I_{1}}}.
\end{cases}
\end{equation}

\textbf{Case 3: }$\lambda_1= 0, \lambda_2\neq 0$

When $\lambda_1=0, \lambda_2\neq0$, the \eqref{eq61} means that

\begin{equation}
\label{eq67}
\begin{cases}
\tau^*=\frac{\lambda_{2}P_{cm}t_{cm}}{\rho\eta\nabla{F^*}-\lambda_{2}E_{tr}},\\
K^*=\frac{1}{\rho\eta\nabla{F^*}-\lambda_{2}E_{tr}}.
\end{cases}
\end{equation}

According to \eqref{eq61}, we have

\begin{equation}
\label{eq68}
P_{cm}t_{cm}K +E_{tr}K\tau = E_{tot}
\end{equation}

Then, we substitute the $\tau^*$ and $K^*$ in \eqref{eq68} and obtain

\begin{equation}
\label{eq69}
\begin{aligned}
\frac{P_{cm}t_{cm}}{\rho\eta\nabla{F^*}-\lambda_{2}E_{tr}} +\frac{\lambda_{2}P_{cm}t_{cm}E_{tr}}{(\rho\eta\nabla{F^*}-\lambda_{2}E_{tr})^2} = E_{tot}.
\end{aligned}
\end{equation}

By solving \eqref{eq69}, we can get $\lambda^*_2=\frac{1}{E_{tr}}(\rho\eta\nabla{F^*}-\sqrt{\frac{\rho\eta\nabla{F^*}P_{cm}t_{cm}}{E_{tot}}})$.

Then by substituting $\lambda_2$ into \eqref{eq69}, we get

\begin{equation}
\label{eq70}
\begin{cases}
\tau_2^*=\frac{1}{E_{tr}}(\sqrt{\rho\eta\nabla{F^*}P_{cm}t_{cm}E_{tot}}-P_{cm}t_{cm}), \\
K_2^*=\sqrt{\frac{E_{tot}}{\rho\eta\nabla{F^*}P_{cm}t_{cm}}}.
\end{cases}
\end{equation}

\textbf{Case 4: }$\lambda_1\neq0, \lambda_2=0$.

When $\lambda_1\neq 0, \lambda_2=0$, the \eqref{eq61} means that

\begin{equation}
\label{eq71}
t_{cm}K+(I_{0}N\mu+a)K\tau^2 = t_{tot}
\end{equation}

Let $\Delta=(\rho\eta\nabla{F^*})^2-4t_{cm}(I_{0}N\mu+a)\lambda^2_1$, and we substitute them into \eqref{eq71}, we get $\tau^*=\frac{\rho\eta\nabla{F^*}\pm\sqrt{\Delta}}{2\lambda_1(I_{0}N\mu+a)}$. However, when $\tau^*=\frac{\rho\eta\nabla{F^*}+\sqrt{\Delta}}{2\lambda_1(I_{0}N\mu+a)}$, $K^*<0$. It is also not realistic.

Thus, we have

\begin{equation}
\label{eq72}
\begin{cases}
\tau^*=\frac{\rho\eta\nabla{F^*}-\sqrt{\Delta}}{2\lambda_1(I_{0}N\mu+a)}, \\
K^*=\frac{1}{\sqrt{\Delta}}.
\end{cases}
\end{equation}

Then, we substitute the $\tau^*$ and $K^*$ in \eqref{eq71} and obtain

\begin{equation}
\label{eq73}
\begin{aligned}
\frac{t_{cm}}{\sqrt{\Delta}}+\frac{(I_{0}N\mu+a)}{\sqrt{\Delta}}\cdot \frac{(\rho\eta\nabla{F^*}-\sqrt{\Delta})^2}{4\lambda^2_1(I_{0}N\mu+a)^2}=E_{tot}.
\end{aligned}
\end{equation}

By solving \eqref{eq73}, we know that $\tau^*<0$, which is also not realistic.

Finally, because $T_i^*=K_i^*\tau_i^*, i=1,2$, to make sure the optimal $\tau$ and $K$ are realistic, which should satisfy the following constraints

\begin{equation}
\label{eq74}
\begin{cases}
\tau^*_1,K^*_1\geq 1,\\
\tau^*_2,K^*_2\geq 1.
\end{cases}
\end{equation}
After some manipulations, we can arrive at \eqref{eq19}. This completes the proof.

}

\bibliographystyle{ieeetr}
\bibliography{ref}

\vfill

\end{document}